%% file: neurips_2019.tex
\pdfoutput=1
\documentclass{article}

\usepackage[preprint]{neurips_2021}
\input{math_commands.tex}

\usepackage{adjustbox} 
\usepackage[dvipsnames]{xcolor}
\usepackage{times}
\usepackage{latexsym}
\usepackage{algorithm}
\usepackage{dsfont}
\usepackage{pgfplots}
\usepackage{subfig}
\usepackage{url}
\usepackage{booktabs}
\usepackage{multirow}
\usepackage[noend]{algpseudocode}
\usepackage{amsmath,amssymb,amsfonts}
\usepackage{bbm}
\usepackage{balance}
\usepackage{lipsum}
\usepackage{graphicx}
\usepackage{mwe}
\usepackage{sidecap}
\usepackage{wrapfig}
\usepackage{amsthm}
\usepackage{namedtensor}

\definecolor{light-gray}{gray}{0.95}

\usepackage[flushleft]{threeparttable}
\usepackage{pifont}% http://ctan.org/pkg/pifont

\newtheorem*{assumption*}{Assumption}

\providecommand{\ie}{\emph{i.e.,} }
\providecommand{\eg}{\emph{e.g.,} }
\providecommand{\parab}[1]{\noindent\textbf{#1}}

\providecommand{\E}{\mathrm{E}}
\providecommand{\parab}[1]{\noindent\textbf{#1}}

\usepackage{lipsum}% http://ctan.org/pkg/lipsum
\usepackage{mdframed}% http://ctan.org/pkg/mdframed
\usepackage{xcolor}% http://ctan.org/pkg/xcolor
\usepackage{pifont}% http://ctan.org/pkg/pifont
\usepackage[utf8]{inputenc} % allow utf-8 input
\usepackage[T1]{fontenc}    % use 8-bit T1 fonts
\usepackage{hyperref}       % hyperlinks
\usepackage{url}            % simple URL typesetting
\usepackage{booktabs}       % professional-quality tables
\usepackage{amsfonts}       % blackboard math symbols
\usepackage{nicefrac}       % compact symbols for 1/2, etc.
\usepackage{microtype}      % microtypography
\usepackage{enumitem}
\usepackage{titlesec}

\newcommand\datasetname{\textsc{HazardWorld}}
\newcommand\modelname{POLCO}

\title{Safe Reinforcement Learning \\ with Natural Language Constraints}

{\centering
\author{Tsung-Yen Yang\thanks{\, Equal contribution.} \\
Princeton University \\
\texttt{ty3@princeton.edu} \\
\And
Michael Hu\footnotemark[1] \\
Princeton University \\
\texttt{myhu@princeton.edu} \\
\And
Yinlam Chow \\
Google Research \\
\texttt{yinlamchow@google.com} \\
\AND
Peter J. Ramadge \\
Princeton University \\
\texttt{ramadge@princeton.edu}\\
\And
Karthik Narasimhan \\
Princeton University \\
\texttt{karthikn@princeton.edu} \\
}}
\begin{document}

\maketitle

\input{abstract}
\input{intro}
\input{related}

\input{formulation}

\input{dataset}
\input{model}
\input{experiment}

\input{conclusion}
\section*{Checklist}

\begin{enumerate}

\item For all authors...
\begin{enumerate}
  \item Do the main claims made in the abstract and introduction accurately reflect the paper's contributions and scope?
    \answerYes{} We show our approach can learn a policy with respect to cost constraints.
  \item Did you describe the limitations of your work?
    \answerYes{} See Section \ref{sec:conlusion}.
  \item Did you discuss any potential negative societal impacts of your work?
    \answerNA{} We do not see any potential negative societal impacts.
  \item Have you read the ethics review guidelines and ensured that your paper conforms to them?
    \answerYes{}
\end{enumerate}

\item If you are including theoretical results...
\begin{enumerate}
  \item Did you state the full set of assumptions of all theoretical results?
    \answerNA{}
	\item Did you include complete proofs of all theoretical results?
    \answerNA{}
\end{enumerate}

\item If you ran experiments...
\begin{enumerate}
  \item Did you include the code, data, and instructions needed to reproduce the main experimental results (either in the supplemental material or as a URL)?
    \answerYes{} Dataset and code to reproduce our experiments are available at~\url{https://sites.google.com/view/polco-hazard-world/}, and see the supplementary material for more details.
  \item Did you specify all the training details (e.g., data splits, hyperparameters, how they were chosen)?
    \answerYes{} See Section \ref{sec:experiments} and the supplementary material.
	\item Did you report error bars (e.g., with respect to the random seed after running experiments multiple times)?
    \answerNA{} We follow the same style of machine learning papers to report the results.
	\item Did you include the total amount of compute and the type of resources used (e.g., type of GPUs, internal cluster, or cloud provider)?
    \answerYes{} See the supplementary material.
\end{enumerate}

\item If you are using existing assets (e.g., code, data, models) or curating/releasing new assets...
\begin{enumerate}
  \item If your work uses existing assets, did you cite the creators?
    \answerYes{} See the supplementary material.
  \item Did you mention the license of the assets?
    \answerYes{} They are open-sourced.
  \item Did you include any new assets either in the supplemental material or as a URL?
    \answerYes{} See the supplementary material.
  \item Did you discuss whether and how consent was obtained from people whose data you're using/curating?
    \answerYes{} We obtained consent to use worker-generated data via Amazon Mechanical Turk.
  \item Did you discuss whether the data you are using/curating contains personally identifiable information or offensive content?
    \answerYes{} We check the data and do not find any identifiable information or offensive content.
\end{enumerate}

\item If you used crowdsourcing or conducted research with human subjects...
\begin{enumerate}
  \item Did you include the full text of instructions given to participants and screenshots, if applicable?
    \answerYes{} See the supplementary material.
  \item Did you describe any potential participant risks, with links to Institutional Review Board (IRB) approvals, if applicable?
    \answerNA{}
  \item Did you include the estimated hourly wage paid to participants and the total amount spent on participant compensation?
    \answerYes{} See the supplementary material.
\end{enumerate}

\end{enumerate}

%%%%%%%%%%%%%%%%%%%%%%%%%%%%%%%%%%%%%%%%%%%%%%%%%%%%%%%%%%%%

% Here is the appendix section
\newpage
\appendix
\input{appendix}

\end{document}

%% file: math_commands.tex
\usepackage{amsmath,amsfonts,bm}

\def\eqref#1{equation~\ref{#1}}
\def\1{\bm{1}}

\DeclareMathAlphabet{\mathsfit}{\encodingdefault}{\sfdefault}{m}{sl}
\SetMathAlphabet{\mathsfit}{bold}{\encodingdefault}{\sfdefault}{bx}{n}

\newcommand{\E}{\mathbb{E}}

\newcommand{\KL}{D_{\mathrm{KL}}}

\DeclareMathOperator*{\argmax}{arg\,max}
\DeclareMathOperator*{\argmin}{arg\,min}

%% file: abstract.tex
% !TEX root = neurips_2019.tex
\begin{abstract}
While safe reinforcement learning (RL) holds great promise for many practical applications like robotics or autonomous cars, current approaches require specifying constraints in mathematical form.
Such specifications demand domain expertise, limiting the adoption of safe RL. In this paper, we propose learning to interpret natural language constraints for safe RL. %
To this end, we first introduce \datasetname{}, a new multi-task benchmark that requires an agent to optimize reward while not violating constraints specified in free-form text.
We then develop an agent with a modular architecture that can interpret and adhere to such textual constraints while learning new tasks. 
Our model consists of \textbf{(1)} a \emph{constraint interpreter} that encodes textual constraints into spatial and temporal representations of forbidden states, 
and \textbf{(2)} a \emph{policy network} that uses these representations to produce a policy achieving minimal constraint violations during training. 
Across different domains in \datasetname{}, we show that our method achieves higher rewards (up to 11x) and fewer constraint violations (by 1.8x) compared to existing approaches.
However, in terms of absolute performance, \datasetname{} still poses significant challenges for agents to learn efficiently, motivating the need for future work.
%
%Dataset and code to reproduce our experiments are available at~\url{https://sites.google.com/view/polco-hazard-world/}.
%
\end{abstract}
%no math, no footnote, no abbreviation in the abstract

%
%{\color{red} Main theme: Few-shot adaptation/bake-in into the physics/training faster]}
% We want to estimate property or the control of the system to optimize some objective functions in the partially observably case.

%% file: intro.tex
\section{Introduction}
\label{sec:intro}
Although reinforcement learning (RL) has shown promise in several simulated domains such as games~\cite{mnih2015dqn,silver2007reinforcement,branavan2012learning} and autonomous navigation~\cite{anderson2018vision,misra2018mapping}, deploying RL in real-world scenarios remains challenging~\cite{dulac2019challenges}. In particular, real-world RL requires ensuring the safety of the agent and its surroundings, which means accounting for \emph{constraints} during training that are orthogonal to maximizing rewards. For example, a cleaning robot must be careful to not knock the television over, even if the television lies on the optimal path to cleaning the house.

\begin{wrapfigure}{R}{0.5\textwidth}
\vspace{-4mm} 
\centering
\includegraphics[width=1.0\linewidth]{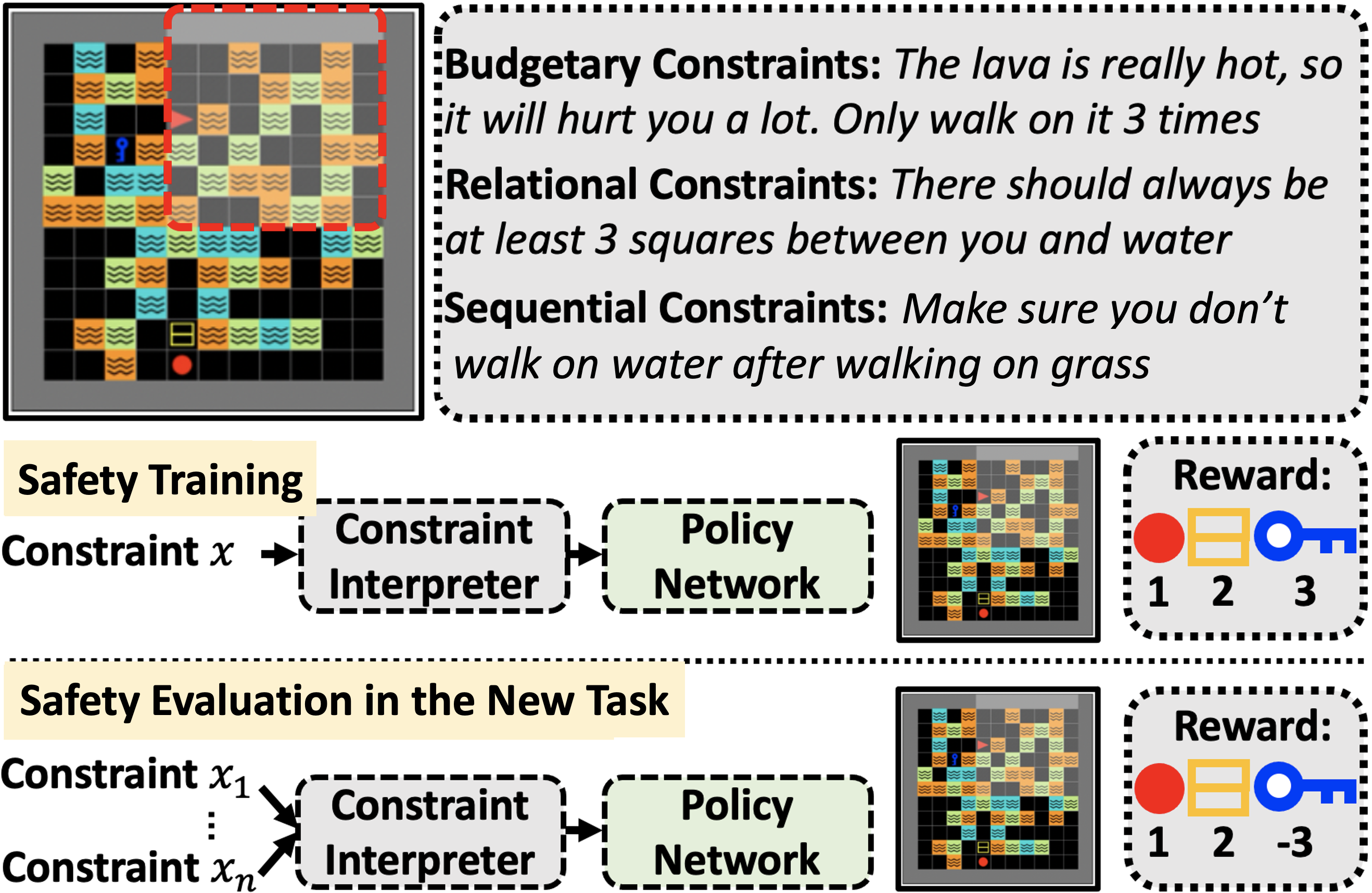}
%\vspace{-5mm} 
\caption{
Learning to navigate with language constraints.
The figure shows 
\textbf{(1)} a third-person view of the environment (red dotted square box),
\textbf{(2)} three types of language constraints,
\textbf{(3)} items which provide rewards when collected.
During safety training, the agent learns to interpret textual constraints while learning the task (\ie collect rewards). During safety evaluation, the agent learns a new task with different rewards while following the constraints and minimizing violations.
}
\label{fig:overview}
\vspace{-0mm}
\end{wrapfigure}

Safe RL tackles these challenges with algorithms that maximize rewards while simultaneously minimizing constraint violations during exploration~\cite{achiam2017constrained,chow2019lyapunov,yang2020projection,yang2020accelerating,Achiam2019BenchmarkingSE,chow2018-srl,berkenkamp2017,elchamie2016-convex-policies,turchetta2020safe,thananjeyan2021recovery}.
However, these algorithms have two key limitations that prevent their widespread use. 
First, they require us to provide constraints in mathematical or logical forms, which calls for specific domain expertise. 
Second, a policy trained with a specific set of constraints cannot be transferred easily to learn new tasks with the same set of constraints, since current approaches do not maintain an explicit notion of constraints separate from reward-maximizing policies. 
This means one would have to retrain the policy (with constraints) from scratch.

We consider the use of \emph{natural language} to specify constraints (which are orthogonal to rewards) on learning. 
Human languages provide an intuitive and easily-accessible medium for describing constraints--not just for machine learning experts or system developers, but also for potential end users interacting with agents such as household robots.
Consider the environment in Fig.~\ref{fig:overview} for example.
Instead of expressing a constraint as
$
 \sum_{t=0}^{T} \1_{s_t\in\mathrm{lava}}\cdot\1_{\not\exists s_{t'}\in\mathrm{water},~t'\in[0,1,...,t-1]}= 0,
$
one could simply say ``\textit{Do not visit the lava before visiting the water}”. The challenge of course, lies in training the RL agent to accurately interpret and adhere to the textual constraints as it learns a policy for the task.
To study this problem, we create \datasetname{}, a collection of grid-world and robotics environments for safe RL with textual constraints (Fig. \ref{fig:overview}). \datasetname{} consists of separate `\textit{safety training}' and `\textit{safety evaluation}' sets, with disjoint sets of reward functions and textual constraints between training and evaluation. To do well on \datasetname{}, an agent has to learn to interpret textual constraints during safety training and safely adhere to any provided constraints while picking up new tasks during the safety evaluation phase. 
Built on existing RL software frameworks~\cite{chevalier2018babyai,Ray2019}, \datasetname{} consists of navigation and object collection tasks with diverse, crowdsourced, free-form text specifying three kinds of constraints: \textbf{(1)
}\emph{budgetary} constraints that limit the frequency of being in unsafe states, \textbf{(2)} \emph{relational} constraints that specify unsafe states in relation to surrounding entities, and \textbf{(3)} \emph{sequential} constraints that activate certain states to be unsafe based on past events (\eg ``\textit{Make sure you don't walk on water after walking on grass}'').
Our setup differs from instruction following~\cite{macmahon2006walk,chen2011learning,artzi2013weakly,misra2017mapping,hermann2020learning,hao2020towards} in two ways. First, instructions specify what to do, while textual constraints only inform the agent on what \textit{not to do}, independent of maximizing rewards. Second, learning textual constraints is a means for ensuring safe exploration while adapting to a new reward function.

In order to demonstrate learning under this setting, we
develop \textit{\textbf{P}olicy \textbf{O}ptimization with \textbf{L}anguage \textbf{CO}nstraints} (\modelname), where we disentangle the representation learning for textual constraints from policy learning. Our model first uses a \emph{constraint interpreter} to encode language constraints into representations of forbidden states. Next, a \emph{policy network} operates on these representations and state observations to produce actions. Factorizing the model in this manner allows the agent to retain its constraint comprehension capabilities while modifying its policy network to learn new tasks. %where natural language constraints are orthogonal to goal instructions.

Our experiments demonstrate that our approach achieves higher rewards (up to 11x) while maintaining lower constraint violations (up to 1.8x) compared to the baselines on two different domains within \datasetname{}. 
Nevertheless, \datasetname{} remains far from being solved, especially in tasks with high-dimensional observations,  complex textual constraints and those requiring high-level planning or memory-based systems.

%% file: related.tex
\section{Related Work}
\label{sec:related}

%

% \begin{figure}[t]
% \centering
% \includegraphics[width=1.01\linewidth]{figure/model}
% \vspace{-8mm}
% \caption{
% %
% \textbf{Model overview.}
% %
% Our model consists of two components.
% %
% \textbf{(1)} The \textit{constraint interpreter} takes a natural language constraint $x$ and an observation $o_t$ as inputs and produces a constraint mask $\hat{M}_C$ and cost constraint threshold prediction $\hat{h}_C$.
% %that encodes locations of the cost entities and the cost constraint threshold $h_C$ that specifies an upper limit of the value of the cost; 
% %
% \textbf{(2)} a \textit{policy network} takes an environment embedding, a constraint mask $\hat{M}_C$, and a cost budget mask $\hat{M}_B$ that specifies cost satisfaction at each step as inputs and produces an embedding.
% %
% We then concatenate the embedding with the embedding of $\hat{h}_C$, followed by an MLP to get an action $a_t.$
% %
% % (Best viewed in color.)
% }
% \label{fig:model}
% \vspace{-5mm}
% %\end{mdframed}
% \end{figure}
\parab{Safe RL.}
Safe RL deals with learning constraint-satisfying policies~\cite{garcia2015comprehensive}, or learning to maximize rewards while minimizing constraint violations.\footnote{In this paper, we consider minimizing constraint violations in expectation~\cite{yang2020projection} and leave stricter notions such as enforcing zero violations~\cite{simao2021alwayssafe} to future work.}
This is a constrained optimization problem, and thus different from simply assigning negative reward values to unsafe states. Furthermore, large negative reward values for constraint violations can destabilize training and lead to degenerate behavior, such as the agent refusing to move.
In prior work, the agent typically learns policies either by \textbf{(1)} exploring the environment to identify forbidden behaviors ~\cite{achiam2017constrained,tessler2018reward,chow2019lyapunov,yang2020projection,stooke2020responsive},
or \textbf{(2)} using expert demonstration data to recognize safe trajectories~\cite{ross2011reduction,rajeswaran2017learning,gao2018reinforcement,yang2020accelerating}.
All these works require a human to specify the cost constraints in mathematical or logical form, and the learned constraints cannot be easily reused for new learning tasks.
In this work, we design a modular architecture to learn to interpret textual constraints, and demonstrate transfer to new learning tasks.
% , which is easier and more flexible to use when specifying constraints.
%
%This reduces the burden of designing the cost functions.
%
%In addition, the policy trained by current safe RL algorithms fail to produce safe actions in the setting where the constraints are changed during run-time.
%
%This means that a single policy can only work for a single constraint.
%
%In contrast, the proposed framework aims to produce safe actions given all types of constraints.

\parab{Instruction following.} 
Our work closely relates to the paradigm of instruction following in RL, which has previously been explored in several environments~\cite{macmahon2006walk,vogel2010learning,chen2011learning,tellex2011understanding,artzi2013weakly,kim2013adapting,Andreas15Instructions, thomason2020vision,luketina2019survey,tellex2020robots,wang2020learning}.
Prior work has also focused on creating realistic vision-language navigation datasets
% with visual urban or household environment
~\cite{bisk2018learning,chen2019touchdown,anderson2018vision, de2018talk}
and proposed computational models to learn multi-modal representations that fuse images with goal instructions~\cite{janner2017representation,blukis2018mapping,fried2018speaker,liu2019representation,jain2019stay,gaddy2019pre,hristov2019disentangled,fu2019language,venkatesh2020spatial}.
Our work differs from the traditional instruction following setup in two ways: \textbf{(1)} Instruction following seeks to (roughly) `translate' an instruction directly into an action policy. This does not apply to our setting since the textual constraints only tell an agent what \textit{not to do}. To actually obtain rewards, the agent has to explore and figure out optimal policies on its own. \textbf{(2)} Since constraints are decoupled from rewards and policies, agents trained to understand certain constraints can transfer their understanding to respect these constraints in new tasks, even when the new optimal policy is drastically different. Therefore, we view this work as orthogonal to traditional instruction following--one could of course combine both instructions and textual constraints to simultaneously advise an agent what to do and what not to do.

% Importantly, our setup is different from instruction following in several ways.
%
% \textbf{(1) Tasks.} Prior approaches in instruction following cannot be applied in our setting since we want to learn a policy that simultaneously optimizes two objectives: rewards and cost constraint satisfaction.
%
% \textbf{(2) Complexity.} In addition, our setting is more challenging since the agent needs to come up with a strategy to avoid cost violations under the constraint budget (\ie budgetary constraints) and the history of trajectories (\ie sequential constraints).
%
% This results in longer planning horizons and requires to understand temporal signals in textual constraints.
%
% In contrast, prior work instruction does not have such complexity.
%

%
%\kn{Contrast more explicitly with LANI and CHAI.}

%
\parab{Constraints in natural language.}
Our notion of `constraints' in this paper differs from prior work that uses instructions to induce planning constraints~\cite{tellex2011understanding,howard2014natural,wang2020learning}--these works again treat instructions as telling the agent how to perform the task. Perhaps closest to this paper is the work of Misra et al.~\cite{misra2018mapping}, which proposes datasets to study spatial and temporal reasoning, containing a subset focusing on \textit{trajectory constraints} (\eg ``\textit{go past the house by the right side of the apple}''). However, they do not disentangle the rewards from the constraints, which may be orthogonal to each other. Prakash et al.~\cite{prakash2020guiding} train a constraint checker to identify whether a constraint (specified in text) has been violated in a trajectory. While their motivation is similar, they ultimately convert constraints to negative rewards, whereas we use a modular approach that allows disentangling reward maximization from minimizing constraint violations and is compatible with modern algorithms for safe RL.

%% file: formulation.tex
\section{Preliminaries}
\label{sec:framework}

% \kn{Can we merge fig 2 with this? They may not understand the constraints before seeing the environment.}
% We now provide a formal description of our problem and task setup.

\parab{Problem formulation.} Our learning problem can be viewed as a partially observable constrained Markov decision process~\cite{altman1999constrained},
which is defined by the tuple $<\mathcal{S},\mathcal{O},\mathcal{A},T,Z,\mathcal{X},R,C>$.
Here $\mathcal{S}$ is the set of states,
$\mathcal{O}$ is the set of observations,
$\mathcal{A}$ is the set of actions, 
$T$ is the conditional probability $T(s'|s,a)$ of the next state $s'$ given the current state $s$ and the action $a,$ and
$Z$ is the conditional probability $Z(o|s)$ of the observation $o$ given the state $s.$
In addition, $\mathcal{X}$ is the set of textual constraint specifications,
$R:\mathcal{S}\times \mathcal{A}\rightarrow \mathbb{R}$ is the reward function, which encodes the immediate reward provided when the agent takes an action $a$ in state $s$,
and
$C:\mathcal{S}\times \mathcal{A}\times \mathcal{X}\rightarrow \mathbb{R}$ is the true underlying constraint function described by $x\in\mathcal{X}$, which specifies positive penalties for constraint violations due to an action $a$ in a state $s$.
Finally, we assume each $x \in \mathcal{X}$ corresponds to a specific cost function $C$. 
%\kn{There seems to be ambiguity here. Is C the set of constraint functions or just one constraint function?}

\parab{RL with constraints.} 
The goal of the learning agent is to acquire a good control policy that maximizes rewards, while adhering to the specified constraints as much as possible during the learning process. Thus, the agent learns a policy $\pi:\mathcal{O}\times\mathcal{X}\rightarrow\mathcal P(\mathcal A)$, which is a mapping from the observation space $\mathcal{O}$ and constraint specification $\mathcal{X}$ 
%\kn{Set X or single x?} 
to the distributions over actions $\cal A$.
%\kn{shouldn’t this be just an action instead of a distribution?}
Let $\gamma\in (0,1)$ denote a discount factor,
$\mu(\mathcal{S})$ denote the initial state distribution,
and $\tau$ denote a trajectory sequence of observations and actions induced by a policy $\pi$, \ie $\tau = (o_0,a_0,o_1,\cdots)$.
For any given $x$, we seek a policy $\pi$ that maximizes the cumulative discounted reward $J_R$ while keeping the cumulative discounted cost $J_C$ below a specified cost constraint threshold $h_C(x)$:%on the sampled trajectory
\[
\max_{\pi}~~J_{R}(\pi)\doteq \mathop{\E}_{\tau\sim\pi}\left[\sum_{t=0}^{\infty}\gamma^{t} R(s_{t},a_{t})\right] \quad
\text{s.t.}~~J_{C}(\pi)\doteq \mathop{\E}_{\tau\sim\pi}\left[\sum_{t=0}^{\infty}\gamma^{t} C(s_{t},a_{t},x)\right]\leq h_C(x),
\]
where $\tau\sim\pi$ is shorthand for indicating that the distribution over trajectories depends on $\pi: s_0\sim\mu, o_{t}\sim Z(\cdot|s_t), a_t\sim\pi(\cdot|o_t,x),s_{t+1}\sim T(\cdot|s_t,a_t)$. 
We use $C(s_t,a_t,x)$ and $h_C(x)$ here to emphasize that both functions depend on the particular constraint specification $x$. 

\parab{Task setup.}
Our goal is to show that constraints specified in natural language allow for generalization to new tasks that require similar constraints during learning. With this in mind, we consider the following safety training and safety evaluation setup:

\parab{(1) Safety training:} During training, we generate random environment layouts and starting states $s_0$ while keeping the reward function $R$ fixed. 
For each episode, we randomly generate a constraint function $C$ and limit $h_C$. We then sample a constraint text $x$ that describes $C$ and $h_C$ from the training set of texts. 
The constraint text $x$ is an input to the agent's policy.
Whenever the agent violates a constraint (at any step), it is provided with a scalar cost penalty learned by the model from $C(s,a,x)$. 
The agent, therefore, sees a variety of different task layouts and constraints, and learns a policy with respect to the constraints for this task as well as how to interpret textual constraints.

\parab{(2) Safety evaluation:} During evaluation, we place the agent in new environments with randomly generated layouts, with a different reward function $R’$. The set of possible constraints $C$ is the same as seen in training, but the corresponding constraint texts are from an unseen test set. During this phase, the agent is not provided any cost penalties from the task. This setup allows us to measure two things: \textbf{(1)} how well an agent can learn new tasks while following previously learned textual constraints, and \textbf{(2)} the applicability of our method when using textual constraints unseen in training.

%% file: dataset.tex
\begin{table*}[t]
\centering
 \subfloat[\datasetname-grid]{
\vspace{0.0in}
 \scalebox{1.0}{
 \begin{tabular}{c|p{9.0cm}}
 \toprule
 \multicolumn{1}{c}{\textbf{Constraint}}
 & \multicolumn{1}{c}{\textbf{Examples}} \\ \hline
 \multirow{1}{*}[-2.5mm]{Budgetary} 
 & \textit{Lava hurts a lot, but you have special shoes that you can use to walk on it, but only up to 5 times, remember!} \\  \hline
 \multirow{1}{*}[0mm]{Relational} 
 & \textit{Water will hurt you if you are two steps or less from them.} \\
 \hline
 \multirow{1}{*}[0mm]{Sequential} 
 & \textit{Make sure you don't walk on water after walking on grass.} \\ 
 \bottomrule
 \end{tabular}}
 \raisebox{-10.0mm}{
\includegraphics[width=0.15\linewidth]{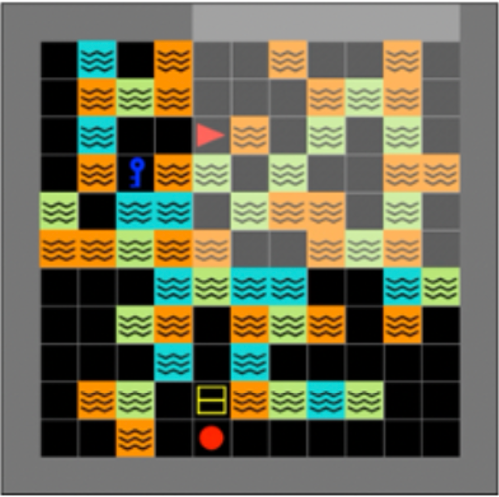}}
 \vspace{+0mm}
 \iffalse
 \raisebox{-3.55mm}{
 \scalebox{0.73}{
 \begin{tabular}{cccc}
 \toprule
 & Count &   Vocab. Size & Mean Length  \\  \midrule
 \multirow{1}{*}{Bud. }      & 432 & 274  & 9.09 $\pm$ 4.30\\
 \multirow{1}{*}{Rel.}      & 262  & 180  & 9.02 $\pm$ 3.65\\
 \multirow{1}{*}{Seq.}      & 290  & 241  & 10.36 $\pm$ 3.49 \\ \hline
 \multirow{1}{*}{Total} & 984 & 526 & 9.44 $\pm$ 3.95 \\
 \bottomrule
 \end{tabular}}}
 \fi
 }
 
 \vspace{-3mm}
 \subfloat[\datasetname-robot]{
  \scalebox{1.0}{
 \begin{tabular}{c|p{9.0cm}}
 \toprule
 \multicolumn{1}{c}{\textbf{Constraint}}
 & \multicolumn{1}{c}{\textbf{Examples}} \\ \hline
 \multirow{1}{*}[-0.3mm]{Budgetary} 
 & \textit{Do not enter the blue square. It is safe to never cross at all.} \\ \hline
  \multirow{1}{*}[-0.3mm]{Relational} 
 & \textit{Three feet is the minimum distance to all the dark blue circles.} \\ \hline
 \multirow{1}{*}[-0.3mm]{Sequential} 
 & \textit{Once a purple box gets touched dark blue circles are disallowed.} \\
 \bottomrule
 \end{tabular}}
 \raisebox{-8mm}{
\includegraphics[width=0.15\linewidth]{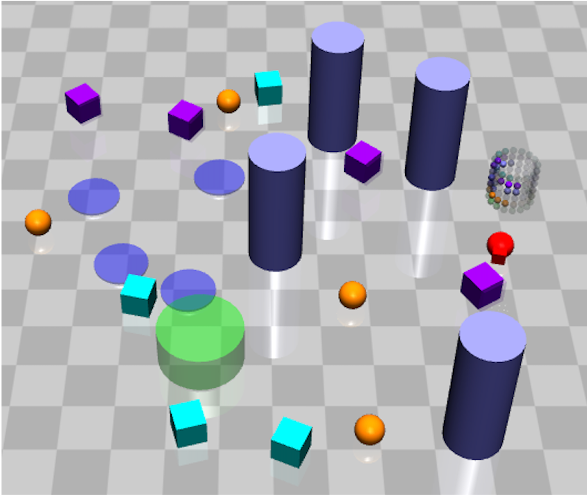}}
 \iffalse
 \raisebox{-3.55mm}{
 \scalebox{1.0}{
 \begin{tabular}{cccc}
 \toprule
 & Count &   Vocab. Size & Mean Length  \\  \midrule
 \multirow{1}{*}{Bud. }      & 1545 & 602  &  8.56 $\pm$ 2.89\\
 \multirow{1}{*}{Seq.}      & 370  & 253  & 9.88 $\pm$ 3.64 \\ \hline
 \multirow{1}{*}{Total} & 1915 & 717 & 8.81 $\pm$ 3.09 \\
 \bottomrule
 \end{tabular}}}
 \fi
 }
 %\vspace{-2mm}
 \caption{Examples of textual constraints for \datasetname-grid and \datasetname-robot. \textbf{(a)} An agent (red triangle) seeks to collect the reward entity (ball, box, key) while avoiding the cost entity (lava, water, grass). \textbf{(b)} An agent (red point) aims to reach a goal position (green area) while avoiding the obstacles (vases, pillars, cubes, \textit{etc}.). Please see the supplementary material for more details.}
 \label{tab:dataset_example}
 \vspace{-4mm}
\end{table*}

\section{\textsc{\datasetname}}
\label{sec:dataset}

\iffalse
\begin{wrapfigure}{R}{0.3\textwidth}
\vspace{-2mm}
\centering
(a)
\raisebox{-0.0mm}{
\includegraphics[width=0.3\linewidth]{figure/lavawall_v2.png}}
(b)
\raisebox{-0.0mm}{
\includegraphics[width=0.35\linewidth]{figure/robotics_v2.png}}
\vspace{-1mm}
    \caption{\datasetname\ 2D and 3D. 
    %
    \textbf{(a)} An agent (red triangle) seeks to collect the reward entity (ball, box, key) while avoiding the cost entity (lava, water, grass). \textbf{(b)} An agent (red point) aims to reach a goal position (green area) while avoiding the obstacles (vases, pillars, cubes, \textit{etc}.).
    }
    \label{fig:env}
    \vspace{-4mm}
\end{wrapfigure}
\fi
To our knowledge, there do not currently exist datasets for evaluating RL agents that obey textual constraints.\footnote{Even though there are several instruction following tasks, our task setup is different, as mentioned previously.}
Thus, we design a new benchmark called \datasetname\ in which the agent starts each episode at a random location within a procedurally generated environment and receives a textual constraint $x$, sampled from a pool of available constraints.
%
% Then we obtain the constraint function $C$ corresponding to this $x$.
%
The agent's goal is to collect all the reward-providing entities while adhering to the specified constraint. Other than the constraint specified, the agent has complete freedom and is not told about how to reach reward-providing states. 

\datasetname\ contains three types of constraints--\textbf{(1)} \textit{budgetary constraints}, which impose a limit on the number of times a set of states can be visited, \textbf{(2)}  \textit{relational constraints}, which define a minimal distance that must be maintained between the agent and a set of entities, and \textbf{(3)} \textit{sequential constraints}, which are constraints that activate unsafe states when a specific condition has been met.
In total, we collect 984 textual constraints for \datasetname-grid (GridWorld environment) and 2,381 textual constraints for \datasetname-robot (robotic tasks).
Table \ref{tab:dataset_example} provides examples.
%
%Please see the supplementary material for more details about the dataset.
%
%During the evaluation phase, we test the performance of the agent using hold-out textual constraints. 
%
%In addition, we further give multiple constraints at a time to the agent which only sees a single constraint during training.
%
%This is to demonstrate that the proposed model can handle unseen combinations of the constraints without retraining the policy as in the prior safe RL algorithms. 
% Appendix~\ref{subsec:appendix_dataset}. 
%

%\kn{For both versions, 2D and 3D in the subsections below, clearly specify the state space, observation space, action space, rewards, constraints and transitions. Might even be worth adding a table. Both subsections below need some restructuring and rewriting.}

\parab{\datasetname-grid.} We implement \datasetname-grid (Table~\ref{tab:dataset_example}(a)) atop the 2D GridWorld layout of BabyAI~\cite{chevalier2018babyai,gym_minigrid}. We randomly place three \textit{reward entities} on the map: `\textit{ball},' `\textit{box},' and `\textit{key},' with rewards of 1, 2, and 3, respectively.
We also randomly place several \textit{cost entities} on the map: `\textit{lava},' `\textit{water},' and `\textit{grass}'. 
We give a cost penalty of 1 when agents step onto any cost entities, which are specified using a textual constraint $x$. The entire state $s_t$ is a grid of size 13$\times$13, including the walls, and the agent's observation $o_t$ is a 7$\times$7 grid of its local surroundings.
There are 4 actions--for moving up, down, left and right.
We use the deterministic transition here.

\parab{Train-test split.} 
We generate two disjoint training and evaluation datasets  $\mathcal{D}_\mathrm{train}$ and $\mathcal{D}_\mathrm{eval}.$
%We train the constraint interpreter
%\kn{Why is training mentioned here? Just say `$\mathcal{D}_\mathrm{train}$ consists of randomly generated maps ....'. Keep it to the point.}
$\mathcal{D}_\mathrm{train}$ consists of 10,000 randomly generated maps paired with 80\% of the textual constraints (787 constraints overall), \ie on average each constraint is paired with 12.70 different maps.
$\mathcal{D}_\mathrm{eval}$ consists of 5,000 randomly generated maps paired with the remaining 20\% of the textual constraints (197 constraints), \ie on average one constraint is paired with 25.38 maps.
In $\mathcal{D}_\mathrm{eval}$ we change the rewards for ball, box, and key to 1, 2, and -3, respectively.
Therefore, in $\mathcal{D}_\mathrm{eval}$, the agent has to avoid collecting the key to maximize reward.
%
%For all the models, we report the results on $\mathcal{D}_\mathrm{eval}.$
%See Appendix \ref{subsec:appendix_parameter} for more details.

% \kn{update this to reflect current procedure.}
%For examples of text and AMT worker prompts, we refer the reader to

%\subsection{{\datasetname}-robot} 
%\label{subsec:safetygym}
\parab{\datasetname-robot.} 
%\kn{This parab is way too long. Make it succinct}
We build \datasetname-robot (Table. \ref{tab:dataset_example}(b)) atop the~\textsc{Safety Gym} environment~\cite{Ray2019} to show the applicability of our model to tasks involving high-dimensional continuous observations.
In this environment, there are five constraint entities paired with textual constraints: \textit{hazards} (dark blue puddles), \textit{vases} (stationary but movable teal cubes), \textit{pillars} (immovable cylinders), \textit{buttons} (touchable orange spheres), and \textit{gremlins} (moving purple cubes).
This task is more challenging than the 2D case since some obstacles are constantly moving.
The agent receives a reward of 4 for reaching a goal position and a cost penalty of 1 for bumping into any constraint entities.
The observation $o_t$ is a vector of size 109, including coordinate location, velocity of the agent, and observations from lidar rays that detect the distance to entities.
%
%The observation also includes simulated sensor values from accelerometers, velocimeters, and gyroscopes.
% action
The  agent has two actions--control signals applied to the actuators  to make it move forward or rotate.
% transition dynamics
The transitions are all deterministic.

\parab{Train-test split.} We follow the same process for obtaining a train-test split as in \datasetname-grid. 
$\mathcal{D}_\mathrm{train}$ consists of 10,000 randomly generated maps paired with 80\% of textual constrains (1,905 constraints), \ie on average one constraint is paired with 5.25 maps.
$\mathcal{D}_\mathrm{eval}$ consists of 1,000 randomly generated maps paired with the remaining 20\% of textual constrains (476 constraints), \ie on average one constraint is paired with 2.10 maps.
In $\mathcal{D}_\mathrm{eval}$ we add four additional goal locations to each map (\ie the maximum reward is 20).
The agent has to learn to navigate to these new locations.

\parab{Data collection.} For the textual constraints in both environments, we collected free-form text in English using Amazon Mechanical Turk (AMT)~\cite{buhrmester2016amazon}. To generate a constraint for \datasetname, we provided workers with a description and picture of the environment, the cost entity to be avoided, and one of the following: \textbf{(a)} the cost budget (budgetary), \textbf{(b)} the minimum safe distance (relational), or \textbf{(c)} the other cost entity impacted by past events (sequential). We then cleaned the collected text by writing a keyword matching script followed by manual verification to ensure the constraints are valid. 

%% file: model.tex
\section{Learning to Interpret Textual Constraints}
\label{sec:model}

\begin{figure*}
\centering
\includegraphics[width=0.63\linewidth]{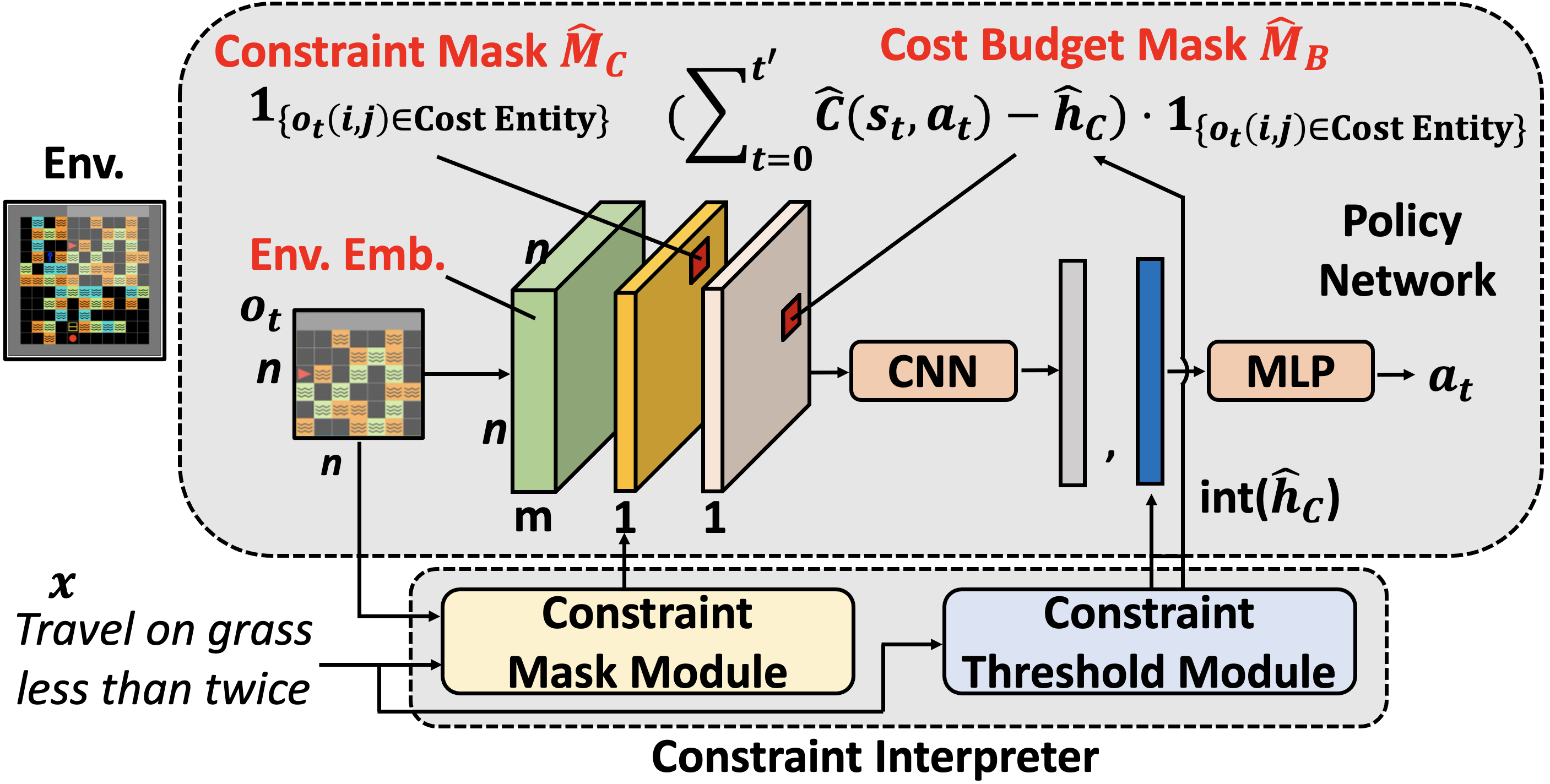}
%\vspace{-2mm}
\caption{
\textbf{Model overview.}
Our model consists of two parts:
\textbf{(1)} the \textit{constraint interpreter} produces a constraint mask and cost constraint threshold prediction from a textual constraint and an observation,
%that encodes locations of the cost entities and the cost constraint threshold $h_C$ that specifies an upper limit of the value of the cost; 
%
\textbf{(2)} a \textit{policy network} takes in these presentations and produces a constraint-satisfying policy.
%
% (Best viewed in color.)
}
\label{fig:model}
%\end{mdframed}
\vspace{-4mm}
\end{figure*}

We seek to train agents that can adhere to textual constraints even when learning policies for new tasks with different reward structures. We now describe our model and training and evaluation procedures.

%\vspace{-0.05in}
\subsection{Model}
We design the RL agent as a deep neural network that consists of two parts (Fig.~\ref{fig:model})--\textbf{(1)} a \textit{constraint interpreter} which processes the text into structured safety criteria (a constraint mask and threshold) and \textbf{(2)} a \textit{policy network} which uses the output of the interpreter along with observations to produce an action.
For simplicity, in the following descriptions, we assume state $s$ and observation $o$ to be 2D matrices, although the model can easily be extended to other input representations.  
%
% (we apply \modelname\ in robotics tasks in Section \ref{sec:experiment}).
%
%In what follows,\\
%Considering a constraint ``\textit{Travel on grass less than twice},''
%
%we need to locate the cost entity on the map (\eg `\textit{grass}') as well as the value of cost constraint threshold (\eg `\textit{two}') in the text constraint.
%
%This allows us to use language to specify the cost constraint in Eq.~(\ref{eq:pcpo_problemFormulation_3}).
%
%Before we describe the proposed model,
%
%\kn{Do we need to repeat these definitions? Aren't they in the previous section? even if we need them, let's make them general, not specific to the dataset.}

\begin{figure*}[t]
%\vspace{-3mm}
\centering
\subfloat[For budgetary and relational constraints]{
\includegraphics[width=0.425\linewidth]{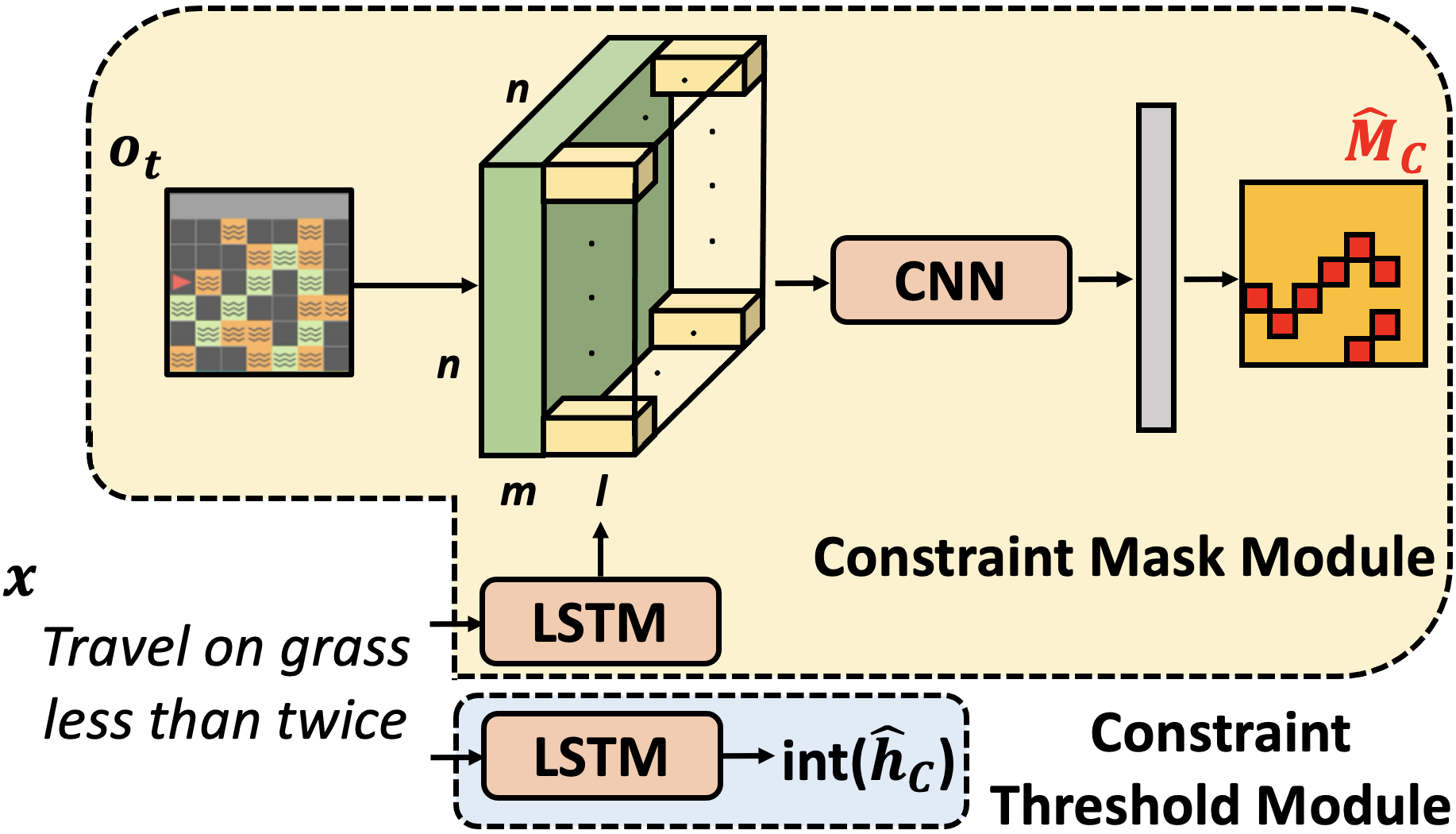}}
\hspace{4mm}
\subfloat[For sequential constraints]{
\includegraphics[width=0.52\linewidth]{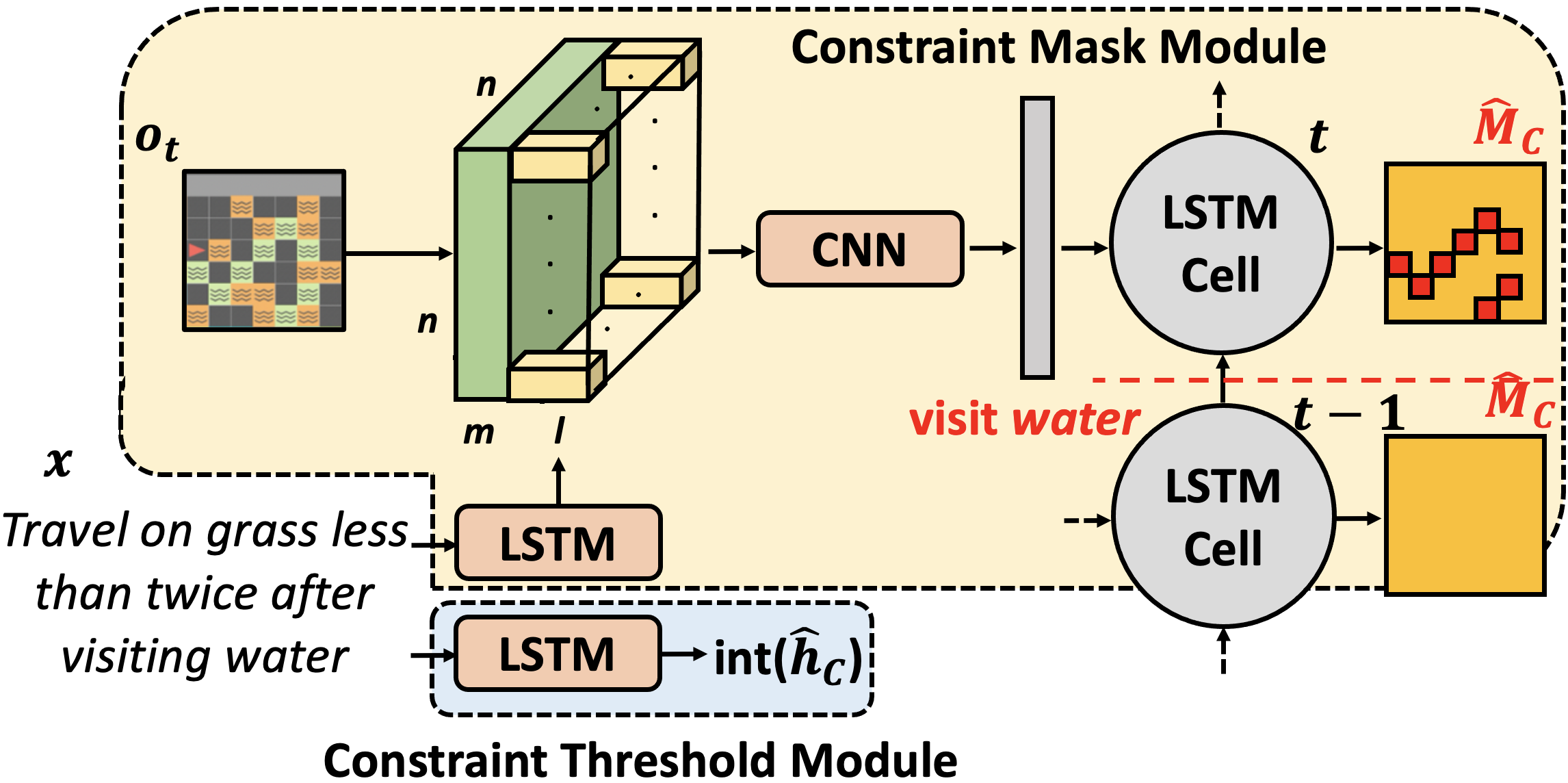}}
\vspace{-2.0mm}
\caption{\textbf{Constraint interpreter.}
\textbf{(a)} For the budgetary and relational constraints, a constraint mask module takes the environment embedding and text vector representation as inputs and predicts $\hat{M}_C.$
\textbf{(b)} For the sequential constraints, we use an LSTM to store the information of the past visited states.
For these three types of constraints, we use another LSTM given $x$ to predict $\hat{h}_C.$
}
\label{fig:model_constraint}
%\end{mdframed}
\vspace{-4mm}
\end{figure*}

\parab{(1) Constraint interpreter (Fig. \ref{fig:model_constraint}).}
We concatenate an observation embedding of size $n\times n\times m$ from observations $o$ of size $n\times n$ with the embedding of the textual constraints $x$ of size $l$ from a long-short-term-memory (LSTM),
followed by using a convolutional neural network (CNN) to get an embedding vector.
We use this vector to produce a constraint mask $\hat{M}_C$,   a binary matrix with the same dimension as $o$--each cell of the matrix is $0/1$ depending on whether the model believes the absence or presence of a constraint-related entity (\eg `\textit{lava}') in the corresponding cell of the observation $o$.
In addition, we feed the textual constraints into an LSTM to produce $\hat{h}_C$, a real-valued scalar which predicts the constraint threshold, \ie the number of times an unsafe state is allowed.

% while $\hat{h}_C$ is a real-valued scalar. $\hat{M}_C$ is a prediction of the true constraint mask, i.e. each cell of the matrix is $0/1$ depending on the absence or presence of a cost entity (e.g. lava) in the corresponding cell of the observation $o$. 

% % The constraint interpreter consists of two parts--a \textit{constraint mask module} and a \textit{constraint threshold module}.

% % %
% % \textbf{(1)} The \textit{constraint mask} module uses the observation $o_t$ and the text $x$ to predict a binary \textit{constraint mask}, denoted by $\hat{M}_C,$ a prediction of the true constraint mask $M_C.$
% % %
% % Each cell in $\hat{M}_C$ is either $1$ if there is a cost entity (\ie the forbidden states are mentioned in the text) in $i$th row and $j$th column of the observation $o_t$ (denoted by $o_t(i,j)$), or zero otherwise. We use $\hat{M}_C$ to identify the cost described in texts while preserving its spatial information.
% % %
% \textbf{(2)} The \textit{constraint threshold} module uses an LSTM to obtain the text vector representation, followed by a dense layer to produce $\hat{h}_C$, a prediction of the true constraint threshold $h_C$.
%

%
For the case of sequential constraints with long-term dependency of the past events, $\hat{M}_C$ will depend on the past states visited by the agent.
For example, in Fig.~\ref{fig:model_constraint}(b), after the agent visits `\textit{water}', $\hat{M}_C$ starts to locate the cost entity (`\textit{grass}').
Thus, for sequential constraints, we modify the interpreter by adding an LSTM layer before computing $\hat{M}_C$ to take the state history into account.
%
%{\color{} For the budgetary and the relational constraints, our model implicitly keeps track of the history using $M_B$, which includes the cumulative cost prediction $\hat{C}_{tot}.$}
%
%This allows us to take advantage of the factorization of the policy network by designing a constraint mask module that producing an appropriate $M_C$ for different types of constraints.
%
Using $\hat{M}_C$ and $\hat{h}_C$ allows us to embed textual constraints in the policy network.

\parab{(2) Policy network.} The policy network produces an action using the state observation $o_t$ and the safety criteria produced by the constraint interpreter.
The environment embedding is concatenated with the constraint mask $\hat{M}_C$ (predicted by the constraint interpreter) and a \textit{cost budget mask}, denoted by $\hat{M}_B$. The cost budget mask is derived from $\hat{h}_C$ (also predicted by the constraint interpreter) and keeps track of the number of constraint violations that the agent has made in the past over the threshold. $\hat{M}_B$ is an $n\times n$ matrix where each element takes the value of $\sum_{t=0}^{t'}\hat{C}(s_t,a_t;x)-\hat{h}_C$ (\ie the value of constraint violations past the budget until $t'$th step) if there is a cost entity in $o_t(i,j),$ or zero otherwise. During the safety evaluation phase, we estimate the cumulative cost $\sum_{t=0}^{t'}\hat{C}(s_t,a_t;x)$ using the predicted $\hat{M}_C$ and the agent's current location at time $t$. After concatenating both the constraint mask $\hat{M}_C$ and cost budget mask $\hat{M}_B$ to the observation embedding, we then feed the resulting tensor into CNN to obtain a vector (grey in Fig.~\ref{fig:model}). This vector is concatenated with a vectorized $\mathrm{int}(\hat{h}_C)$ (\ie $\hat{h}_C$ rounded down) and fed into an MLP to produce an action.

\parab{\modelname\ in \datasetname-robot.}
To apply \modelname\ in this environment, the constraint interpreter predicts the cost entity given the textual constraints.
We then map the cost entity to the pre-defined embedding vector (\ie one-hot encoding).
% %
We then concatenate the embedding vector, the embeddings of the predicted $\hat{h}_C,$ and the value of cost budget (rounded down) to the observation vector.
%This information is further concatenated with the observation of the agent which measures the distance to surrounding objects. 
Finally, the policy network takes in this concatenated observation and produces a safe action.

\parab{Advantages of the design.} The design of \modelname\ tightly incorporates textual constraints into the policy network. Our model factorization--into \textbf{(1)} a constraint interpreter and \textbf{(2)} a policy network--allows us to design specific constraint interpreters for different types of constraints.\footnote{$\hat{M}_B$ equates to a scaled up version of $\hat{M}_C$ since we assume only one constraint specification per episode, but this is not necessary in general since we may have multiple constraints over different cost entities. In that case, $\hat{M}_B$ may have different cost budgets for different cells (entities).} Furthermore, our approach scales gracefully to handling multiple constraints. While existing safe RL algorithms require retraining the policy for each unique combination of constraints, we can simply add together the $\hat{M}_{C}$ of each constraint to handle multiple constraints imposed simultaneously.

\subsection{Safety training}
We first train the constraint interpreter using a random policy to collect trajectories, and then we use the trained interpreter to predict constraints while training the policy network.

%
%For the constraint interpreter, we use a random policy to sample trajectories and learn from any constraint violations. For the policy network, we use a constrained policy optimization algorithm to learn from rewards and the constraint interpreter's predictions.

\parab{Stage 1: Interpreter learning.}
We use a random policy to explore the environment, and obtain trajectories consisting of observations $o_t,$ along with the corresponding textual constraint $x.$
Using the constraint violations encountered in the trajectory and the cost specification $C$, we obtain a target $M_C$ for training the constraint interpreter.
In addition, we also obtain the ground-truth value of $h_C$ for learning the constraint threshold module.

%the queried ground-truth cost values and $h_C$ values, to train the constraint interpreter.
%by using a random policy to obtain trajectories over observations $o_t$ along with corresponding text constraints $x$ from the training set $\mathcal{D}_\mathrm{train}.$ For the observations in $\mathcal{D}_\mathrm{inter}$, the agent query the corresponding ground-truth $M_C$ and $h_C$ values to use as supervision.
%
%Note that even though our approach requires learning the constraint interpreter with a few samples, we demonstrate that integrating language understanding into modern safe RL algorithms is possible and better than simply concatenating the text representation and the observation with end-to-end training.
%
We train the constraint mask module of the constraint interpreter by minimizing the following binary cross-entropy loss over these trajectories:
$\mathcal{L}(\Theta_1)=-\E_{(o_t,x)\sim\mathcal{D}_\mathrm{train}}\Big[\frac{1}{|M_C|}\sum_{i,j=1}^n y\log \hat{y}+(1-y)\log (1-\hat{y})\Big],$
where $y$ is the target $M_C(i,j;o_t,x)$, which denotes the target (binary) mask label in $i$th row and $j$th column of the $n\times n$ observation $o_t,$ $\hat{y}$ is the predicted $\hat{M}_C(i,j;o_t,x)$, \ie the probability prediction of constraint mask, and $\Theta_1$ are the parameters of the constraint mask module.
%$M_C(i,j;o_t,x)$  and $x,$ and $\hat{M}_C(i,j;o_t,x)$ .
%

% %
% \iffalse
%  \begin{algorithm}[t]
%  \centering
%   \caption{Learning algorithm for~\modelname}
%   \label{algo:PCPO}
%   \begin{algorithmic}%[]
%       \State\textbf{Stage 1 (Interpreter learning)}
%       \State \quad Initialize a policy $\pi^0=\pi(\cdot|\vtheta^0)$ and a buffer $\mathcal{B}$
%       \State\quad Run $\pi^0$ and store trajectories in $\mathcal{B}$ 
%       \State\quad Train $\Theta_1$ and  $\Theta_2$ using Eq.~(\ref{eq:sl_1}) and Eq.~(\ref{eq:sl_2})
       
%       \State\textbf{Stage 2 (Policy learning)}
%       \State \quad Empty $\mathcal{B}$
%       \State \quad \textbf{For}$~~k=0,1,2,\cdots~~$\textbf{do}
%              \State \quad\quad\quad Run $\pi^{k}=\pi(\cdot|\vtheta^{k})$ and store trajectories in $\mathcal{B}$
%              %\If{not an $\epsilon$-FOSP}
%                  \State \quad\quad\quad Update $\vtheta^{k+1}$ using any safe RL algorithms
%              \State \quad\quad \quad Empty $\mathcal{B}$

%       %\State \textbf{return} Optimal $\vtheta^{*}$
%      %\EndProcedure
%   \end{algorithmic} 
% \end{algorithm}
% %\vspace{-2mm}
% \fi

For the constraint threshold module, we minimize the following loss:
$\mathcal{L}(\Theta_2)=\E_{(o_t,x)\sim\mathcal{D}_\mathrm{train}}\big[(h_C(x)-\hat{h}_C(x))^2\big],$
where $\Theta_2$ are the parameters of the constraint threshold module.

This approach ensures cost satisfaction \textit{during both policy learning and safety evaluation}, an important feature of safe RL. If we train both the policy and the interpreter simultaneously, then we risk optimizing according to inaccurate $\hat{M}_C$ and $\hat{h}_C$ values, as observed in our experiments.

\parab{Stage 2: Policy learning.} We  use a safe RL algorithm called projection-based constrained policy optimization (PCPO) \cite{yang2020projection} to train the policy network.
During training, the agent interacts with the environment to obtain rewards and penalty costs ($\hat{M}_C$) are provided from the trained constraint interpreter for computing $J_R(\pi)$ and $J_C(\pi)$ (ground-truth $C$ is not used).
%
% We briefly review PCPO here.
%
PCPO is an iterative method that performs two key steps in each iteration\footnote{One can use other safe RL algorithms such as Constrained Policy Optimization (CPO) \cite{achiam2017constrained}}--optimize the policy according to reward and project the policy to a set of policies that satisfy the constraint. 
During safety evaluation, we evaluate our model in the new task with the new reward function and the textual constraints from $\mathcal{D}_\mathrm{eval}.$

\subsection{Safety evaluation}
% We perform two types of safety evaluations:

\textbf{(1) Transfer to new tasks:}
We take the policy trained in $\mathcal{D}_\mathrm{train}$ and fine-tune it on tasks having new reward functions with textual constraints from $\mathcal{D}_\mathrm{eval}.$
We \textit{do not} retrain the constraint interpreter on $\mathcal{D}_\mathrm{eval}.$
The policy is fine-tuned to complete the new tasks without the penalty signals from the cost function $C.$
In \datasetname-robot, we optimize the policy using CPO~\cite{achiam2017constrained}.% since it has better performance than PCPO. \\
\textbf{(2) Handling multiple textual constraints:}
We also test the ability of our model to handle multiple constraints imposed simultaneously (from $\mathcal{D}_\mathrm{eval}$), by adding the cost constraint masks $\hat{M}_C$ of each constraint together when given multiple constraints.
During safety training, the policy is still trained with a single constraint. No fine-tuning is performed and the reward function is maintained the same across training and evaluation in this case.
%
% During safety evaluation, the policy is evaluated on multiple constraints (from $\mathcal{D}_\mathrm{eval}$) imposed at the same time ().
%
% The reward function is the same for $\mathcal{D}_\mathrm{train}$ and  $\mathcal{D}_\mathrm{eval}.$

%% file: experiment.tex
%\vspace{-0.1in}

\section{Experiments}
\label{sec:experiments}
Our experiments aim to study the following questions: \textbf{(1)} Does the policy network, using representations from the constraint interpreter, achieve fewer constraint violations in new tasks with different reward functions?
\textbf{(2)} How does each component in \modelname\ affect its performance?

\subsection{Setup}

\parab{Baselines.} 
We consider the following baselines: \\
\textbf{(1)}  \textit{Constraint-Fusion (CF) with PCPO}: This model~\cite{walsman2019early} takes a concatenation of the observations and text representations as inputs (without $M_C,$ $M_B$ and $h_C$) and produces an action, trained with an end-to-end approach using PCPO.
%
%This is similar to the approach of FiLM \cite{dumoulin2018feature-wise}, in which we fuse both representations. \mh{Is this true? We don't learn any scaling operations here, so it's less expressive than FiLM}
%
This model jointly processes the observations and the constraints. \\
\textbf{(2)}\textit{ CF with TRPO:}
We train CF using trust region policy optimization (TRPO) ~\cite{schulman2015trust}, which \textit{ignores} all constraints and only optimizes the reward.
This is to demonstrate that the agent will have substantial constraint violations when ignoring constraints.\\
\textbf{(3)} \textit{Random Walk (RW)}: We also include a random walk (RW) baseline, where the agent samples actions uniformly at random.

\parab{Evaluation metrics.}
To evaluate models, we use \textbf{(1)} the average value of the reward $J_R(\pi)$,
and \textbf{(2)} the average constraint violations $\Delta_C:=\max(0,J_C(\pi)-h_C).$
Good models should have a small $\Delta_C$ (\ie close to zero) while maximizing $J_R(\pi).$
More details on the implementation, hyper-parameters, and computational resources are included in the supplementary material.

\subsection{Results}

\begin{figure*}[t]
\captionsetup[subfloat]{justification=centering}
\centering
%\subfloat[]{
(a) 
\raisebox{-2.0mm}{
\includegraphics[width=0.47\linewidth]{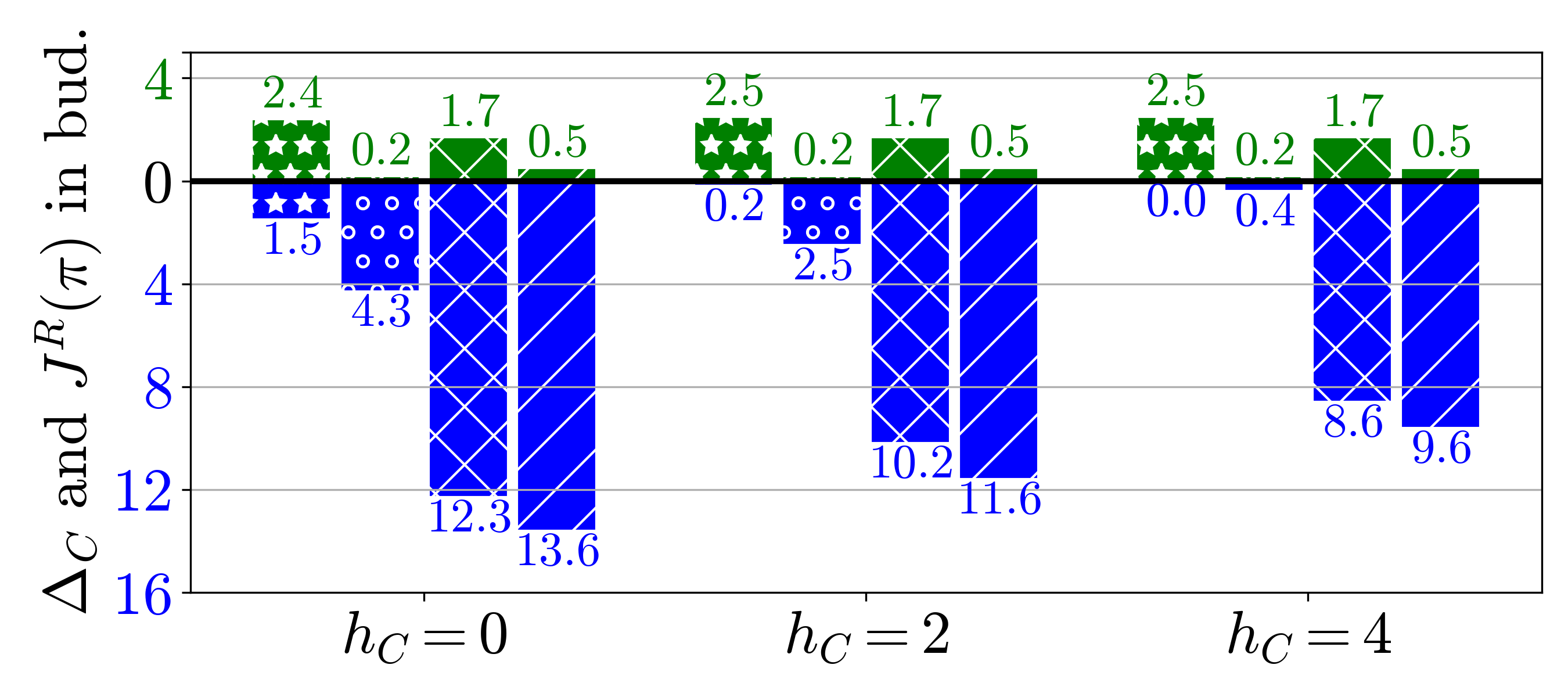}}
\raisebox{-2.0mm}{
\includegraphics[width=0.154\linewidth]{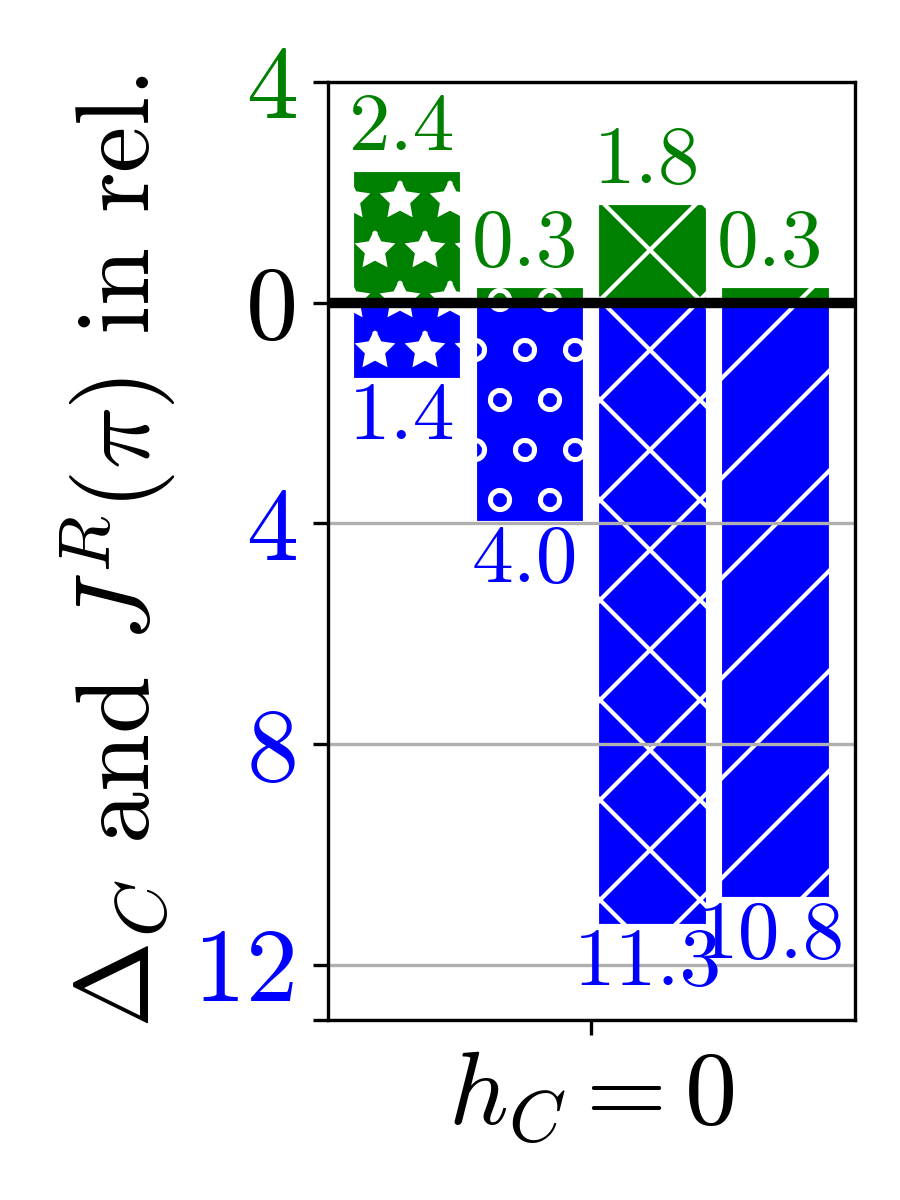}}
\raisebox{-2.0mm}{
\includegraphics[width=0.154\linewidth]{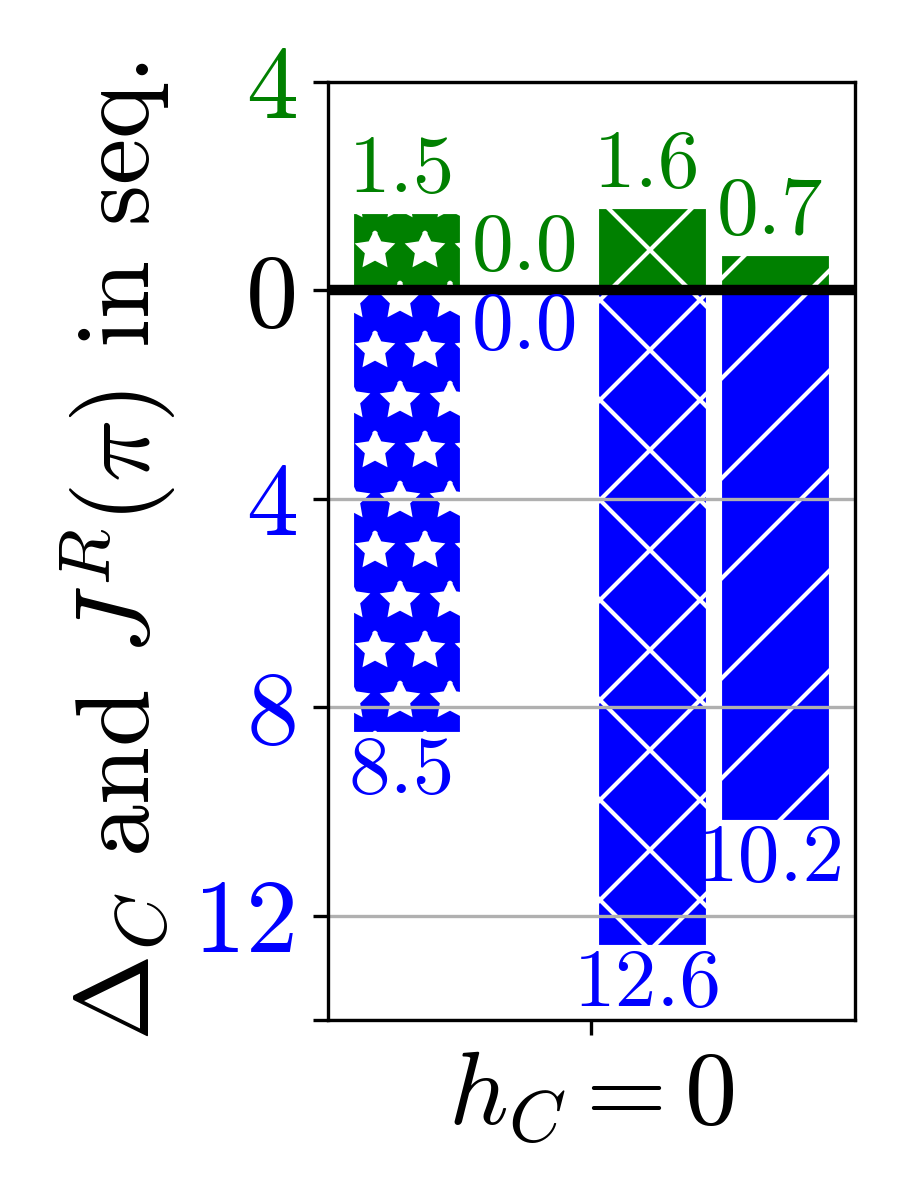}}
%}

\vspace{-0.5mm}

%\subfloat[Results for handling multiple language constraints in \datasetname\ 2D]{
(b)
\raisebox{-2.0mm}{
\includegraphics[width=0.47\linewidth]{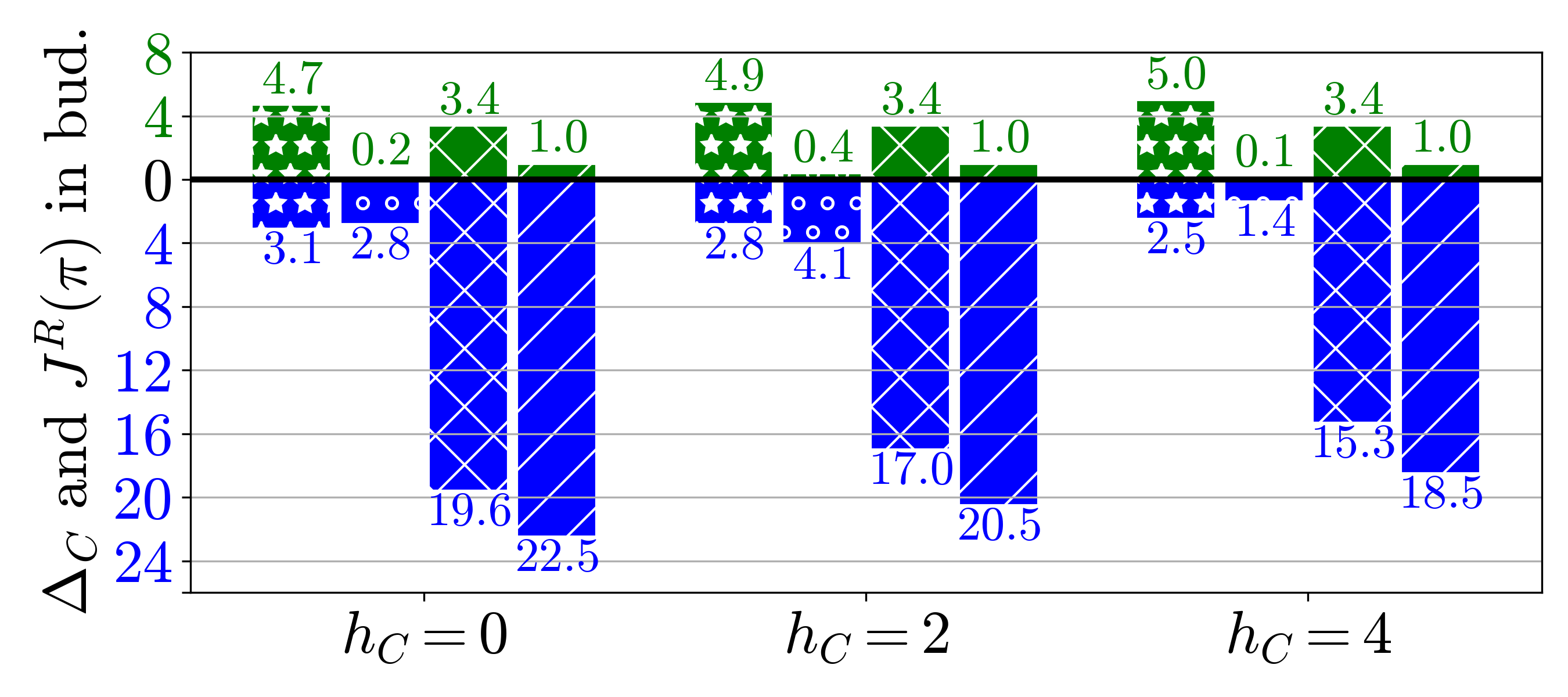}}
\raisebox{-2.0mm}{
\includegraphics[width=0.154\linewidth]{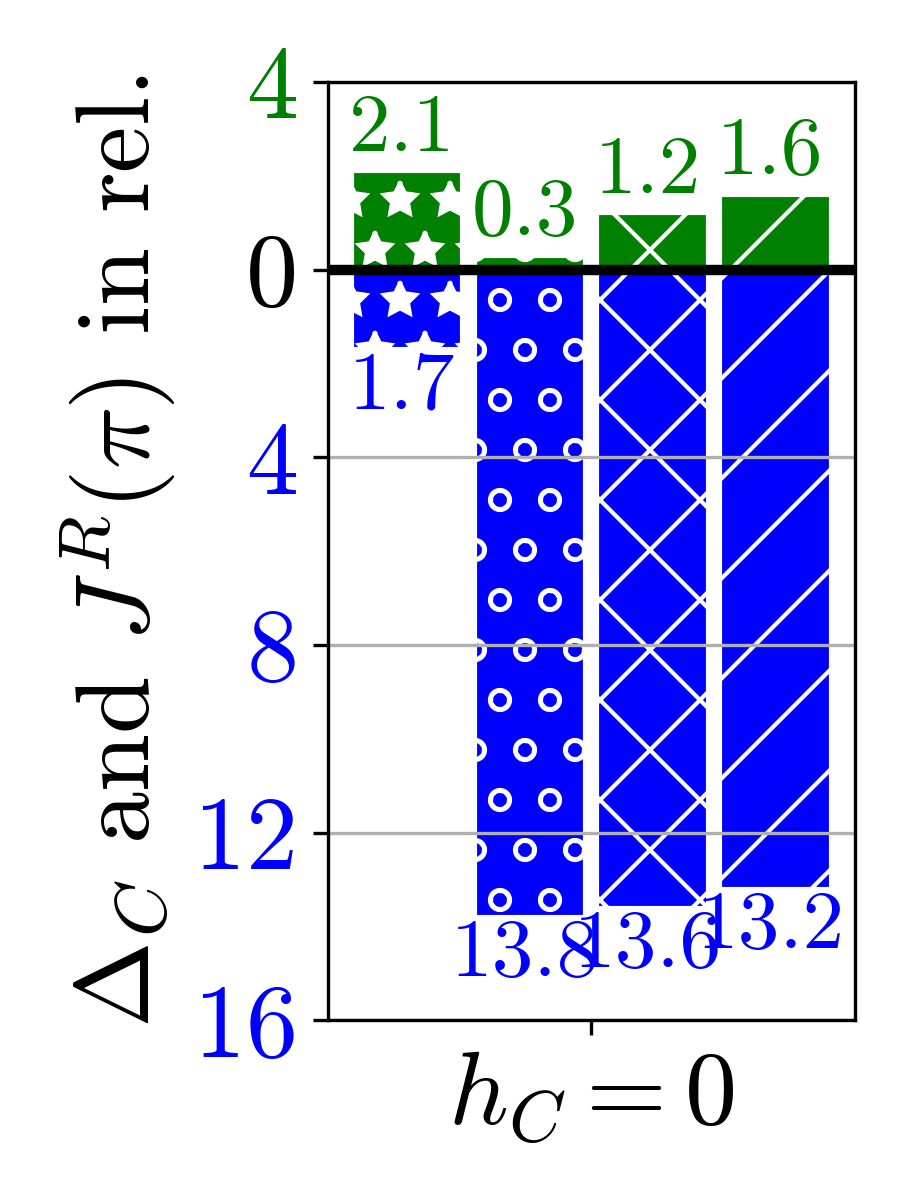}}
\raisebox{-2.0mm}{
\includegraphics[width=0.154\linewidth]{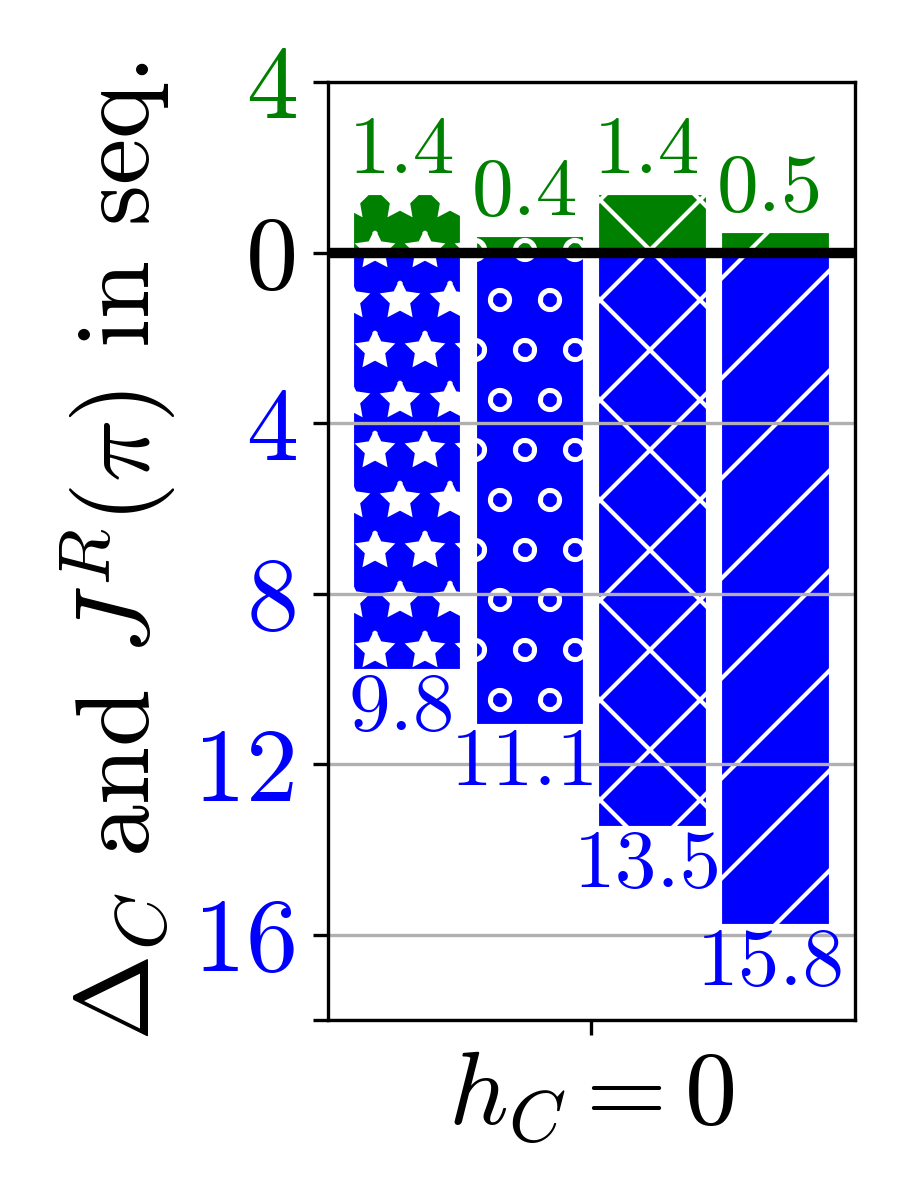}}
%}

\adjustbox{lap={\width}{-1.5mm}}{\raisebox{0.0mm}{
\includegraphics[width=1.0\linewidth]{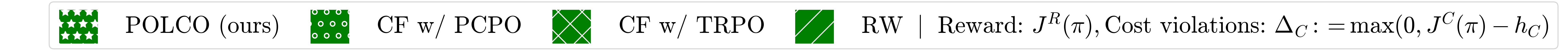}}}
%\vspace{-5mm}
\caption{\label{tab:predictorPerformance}
Results in \datasetname-grid over different values of $h_C.$
These graphs represent the results of budgetary, relational, and sequential constraints, respectively.
The \textbf{\color{OliveGreen}green bars} are the reward performance ($J^R(\pi)$) and the \textbf{{\color{blue}blue bars}} are the constraint violations ($\Delta_C$). 
For $J^R(\pi)$, higher values are better and for $\Delta_C$, lower values are better.
\textbf{(a)}
%
%Performance of the policy fine-tuned in the new reward function than that of in training phase over the tested models and algorithms.
Results for transfer to the new tasks.
\textbf{(b)} 
%Performance of the models with multiple constraints imposed simultaneously during evaluation. 
%
%The policy here is trained only with the single constraint.
Results for handling multiple textual constraints.
\modelname\ generalizes to unseen reward structure and handle multiple constraints with minimal constraint violations in the new task. 
}
\vspace{-3mm}
\end{figure*}

\begin{figure*}[]
\centering
\raisebox{-2.0mm}{
\includegraphics[width=0.47\linewidth]{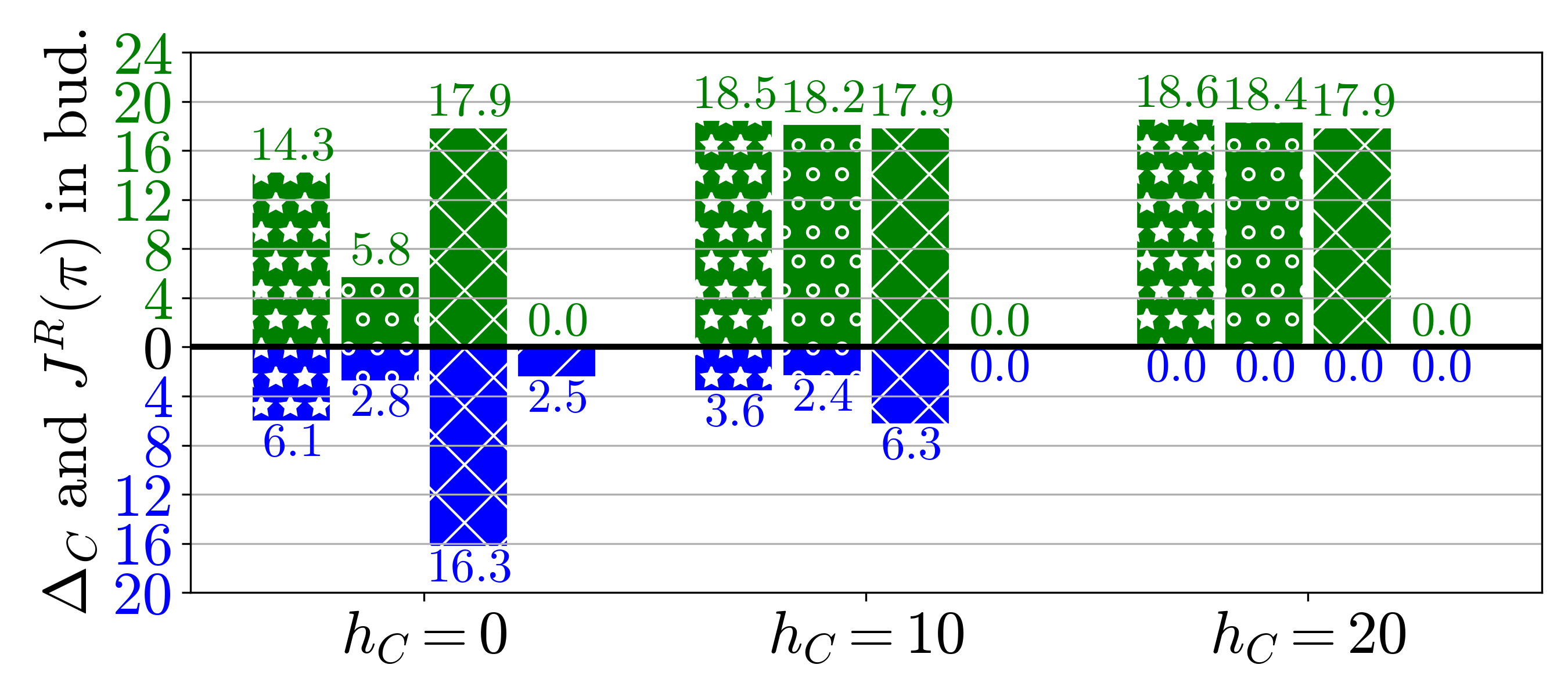}}
\raisebox{-2.0mm}{
\includegraphics[width=0.154\linewidth]{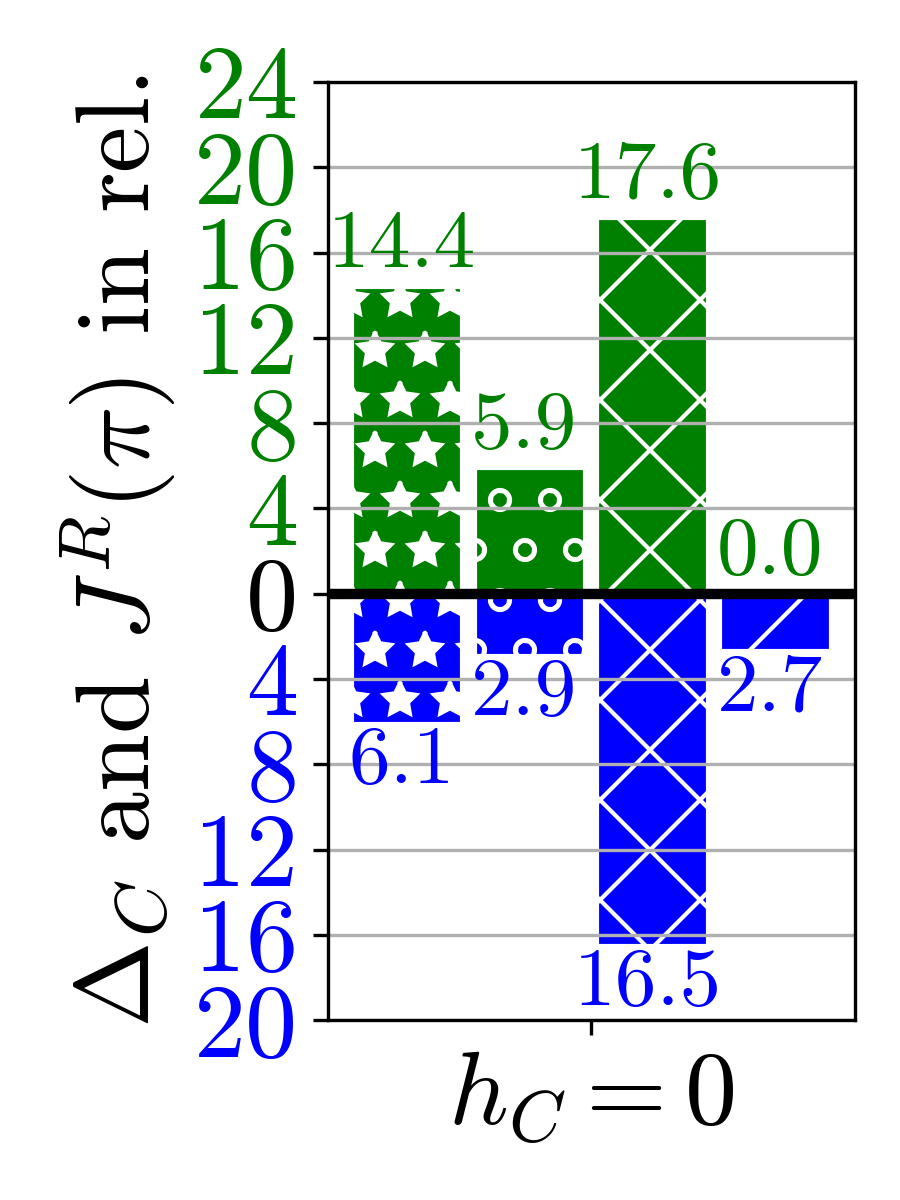}}
\raisebox{-2.0mm}{
\includegraphics[width=0.154\linewidth]{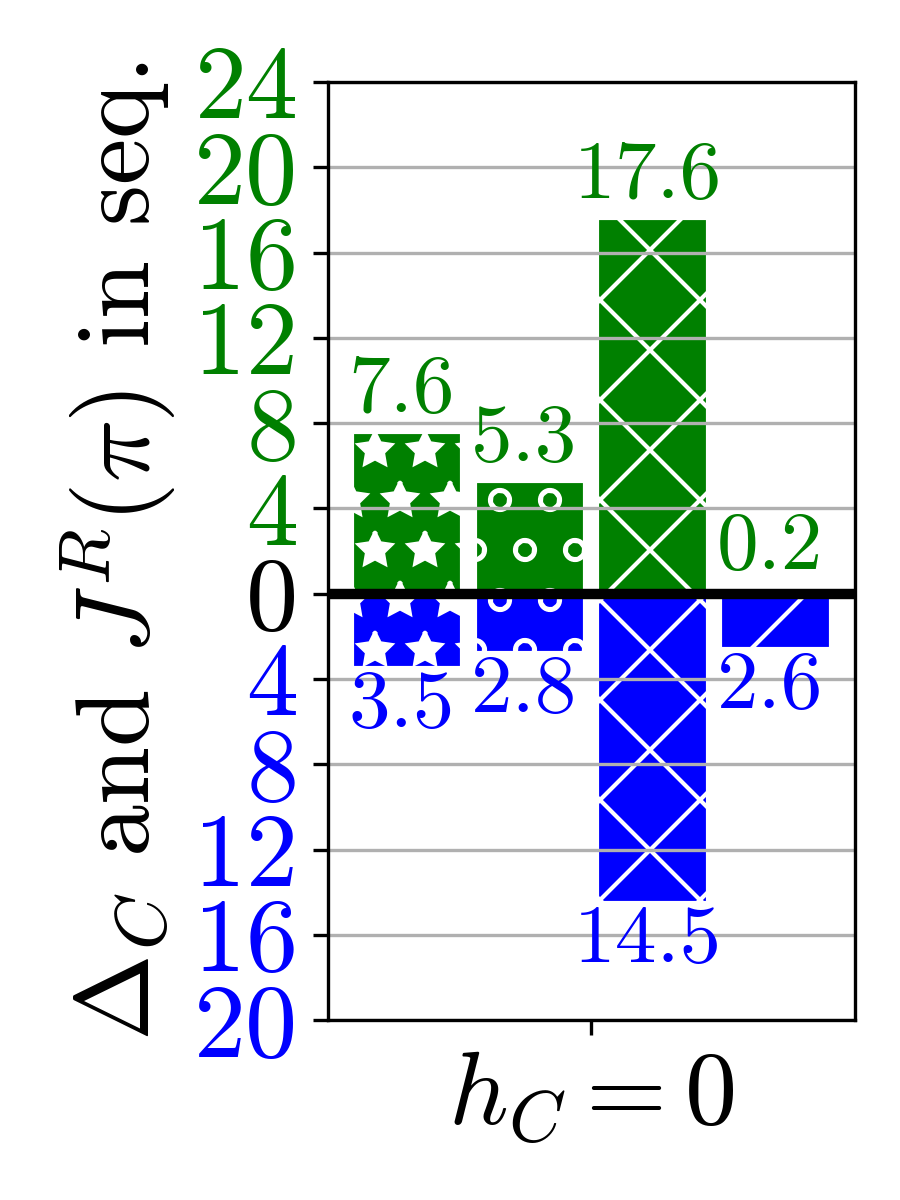}}
\adjustbox{lap={\width}{-1.5mm}}{\raisebox{0.0mm}{
\includegraphics[width=1.0\linewidth]{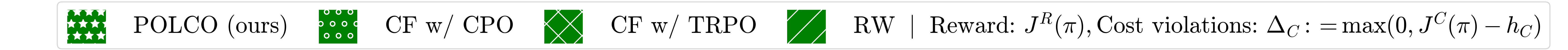}}}
%\vspace{-5mm}
    \caption{
    Results in \datasetname-robot over different values of $h_C$ for transfer to the new tasks.
%
%The \textbf{\color{OliveGreen}green bars} are the reward performance ($J^R(\pi)$) and the \textbf{{\color{blue}blue bars}} are the constraint violations ($\Delta_C$).
%
\modelname\ achieves competitive results with higher rewards and lower cost violations.}
    \label{tab:roboticsTask}
    \vspace{-4mm}
\end{figure*}

%

%
\iffalse
\parab{Training details.}
%
For the policy network, the rollout length of the task and batch size for one policy update are set to $200$ and $50,$ respectively.
%
The step size is set to $10^{-3}$ with $2,500$ iterations for all tested approaches.
%
For the constraint interpreter, we train three of them for each types of constraints.
%
This is to ensure that the constraint interpreter can robustly and accurately produce the cost map.
%
To generate the training set, we use a random policy to generate $10^{6}$ state-action pairs with the constraints sampling from the training set.
%
The batch size for one update is set to $256$.
%
The step size is set to $10^{-3}$ with $10,000$ iterations for all constraint interpreters.
%
%Please read Appendix \ref{subsec:appendix_experiment} for the details of the experiments and the parameters.
\fi

\label{sec:results}
\iffalse
Our experiments study the following five questions: 
%
\textbf{(1)} How does \modelname\ generalize to unseen environments with different reward functions or layouts (\ie can we reuse the constraint interpreter in new settings)?
%
\textbf{(2)} How does \modelname\ trained with a single constraint generalize to multiple constraints imposed simultaneously (\ie can we recombine constraints)?
%
\textbf{(3)} How does \modelname\ perform compared to end-to-end training approaches, where the input is a simple concatenation of the text and the observation?
%
\textbf{(4)} How does each model component--$M_C,$ $M_B$ and $h_C$ affect performance?
%
\textbf{(5)} How does \modelname\ perform in high-dimensional control tasks?
\fi

\parab{\datasetname-grid.}
Fig.~\ref{tab:predictorPerformance}(a) shows results for all models in the first evaluation setting of transfer to new tasks.
\modelname{} has lower constraint violations in excess of $h_C$ while still achieving better reward performance in all cases. In comparison, the high cost values ($\Delta_C$) obtained by RW and CF with TRPO indicate the challenges of task.
This supports our idea of using the learned constraint interpreter to learn a new task with similar textual constraints while ensuring constraint satisfaction.
CF with PCPO has higher constraint violations, and in the most cases, does not optimize the reward, which suggests that it cannot transfer the constraint understanding learned in $\mathcal{D}_\mathrm{train}$ to $\mathcal{D}_\mathrm{eval}.$ 

Fig. \ref{tab:predictorPerformance}(b) shows our evaluation with multiple textual constraints.
We see that \modelname\ achieves superior reward and cost performance compared to the baselines, while CF with PCPO has worse reward and cost performance.
%
%This implies that the design of the policy network is effective in handling multiple constraints unseen during training.
%
% In contrast, 
%
This shows that our approach is flexible enough to impose multiple constraints than that of existing safe RL methods which requires retraining the policy for each unique combination of constraints.
%\kn{is this a valid claim?}
%
%These experiments show that \modelname\ induces a library of constraint representations that can be reused or recombined to achieve lower constraint violations in the unseen environments.
%
%In addition, the design of the policy network makes it easier for safe RL algorithms to transfer cost information to the new tasks and hence optimize.

\parab{\datasetname-robot.}
%
% Finally, we test our model in \datasetname~3D to demonstrate the applicability of our approach.
%
%Results on \datasetname~3D are presented in Fig. \ref{tab:roboticsTask}.
Fig. \ref{tab:roboticsTask} shows transfer to new tasks in \datasetname-robot.
The $J^R(\pi)$ and $\Delta_C$ of RW is relatively small since the agent does not move much because of random force applied to each actuator.
For the budgetary constraints, although CF with TRPO achieves the best reward when $h_C=0,$ it has very large constraint violations.
\modelname\ performs better than the baselines--it induces policies with higher reward under fewer constraint violations in most cases.
In contrast, CF with CPO has lower reward performance.

\iffalse
%
In addition, Table~\ref{tab:predictorPerformance}(b,e,f) shows zero-shot performance (\ie no fine-tuning) over different $h_C$ of agents with multiple constraints imposed at the same time.
%
The policy here only sees a single constraint during training, and the reward is the same during training and zero-shot testing.
%
We observe that in experiments with a larger $h_C,$ generally the agents are less constrained during navigation and yield higher reward and lower constraint violations.
%
Still, in most cases the baseline agents have substantially higher constraint violations or obtain minimal reward.
%
These two experiments show that \modelname\ induces a library of the constraint representations that can be reused or recombined to achieve lower constraint violations in the unseen environments.
\fi

%

\begin{figure*}[t]
\centering
%
%\subfloat[Performance of the models evaluated with the same reward function]{
\raisebox{-2.0mm}{
\includegraphics[width=0.47\linewidth]{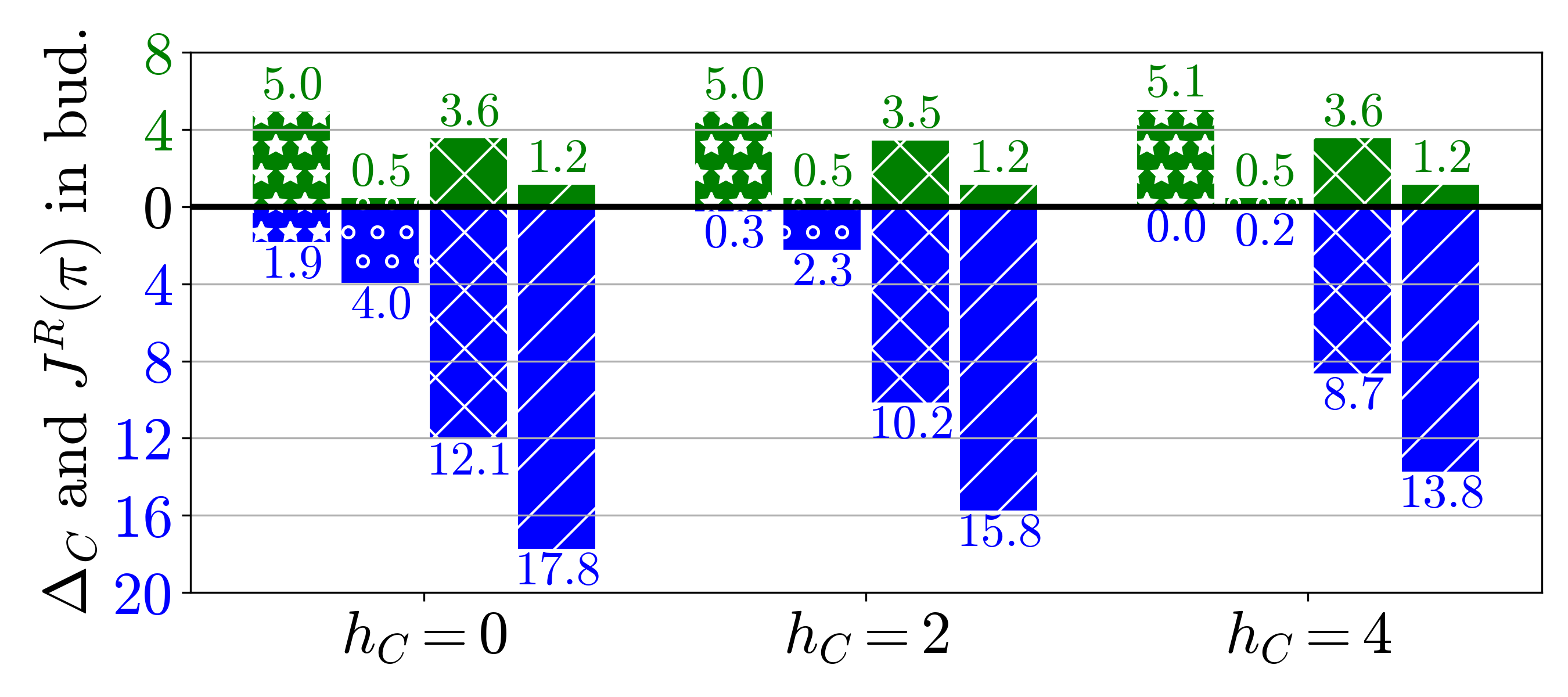}}
\raisebox{-2.0mm}{
\includegraphics[width=0.154\linewidth]{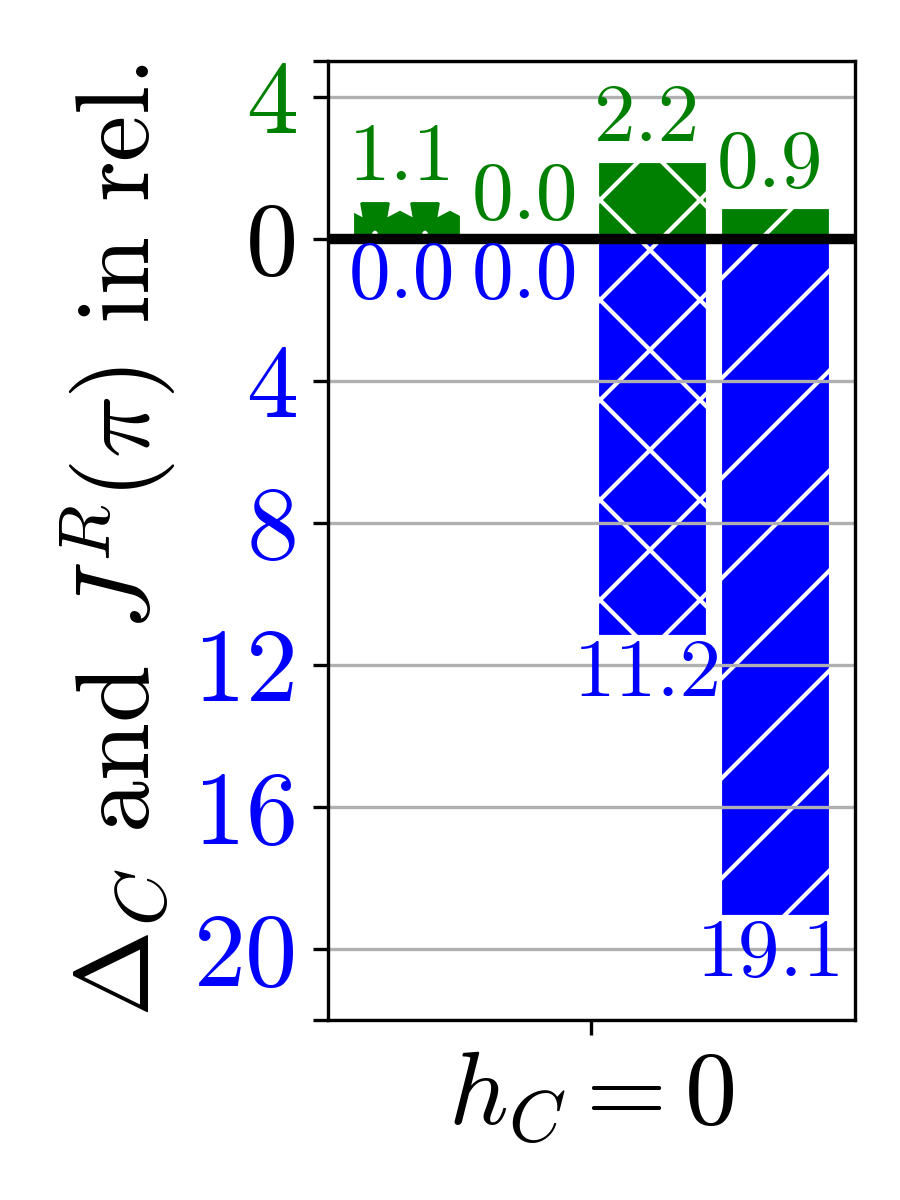}}
\raisebox{-2.0mm}{
\includegraphics[width=0.154\linewidth]{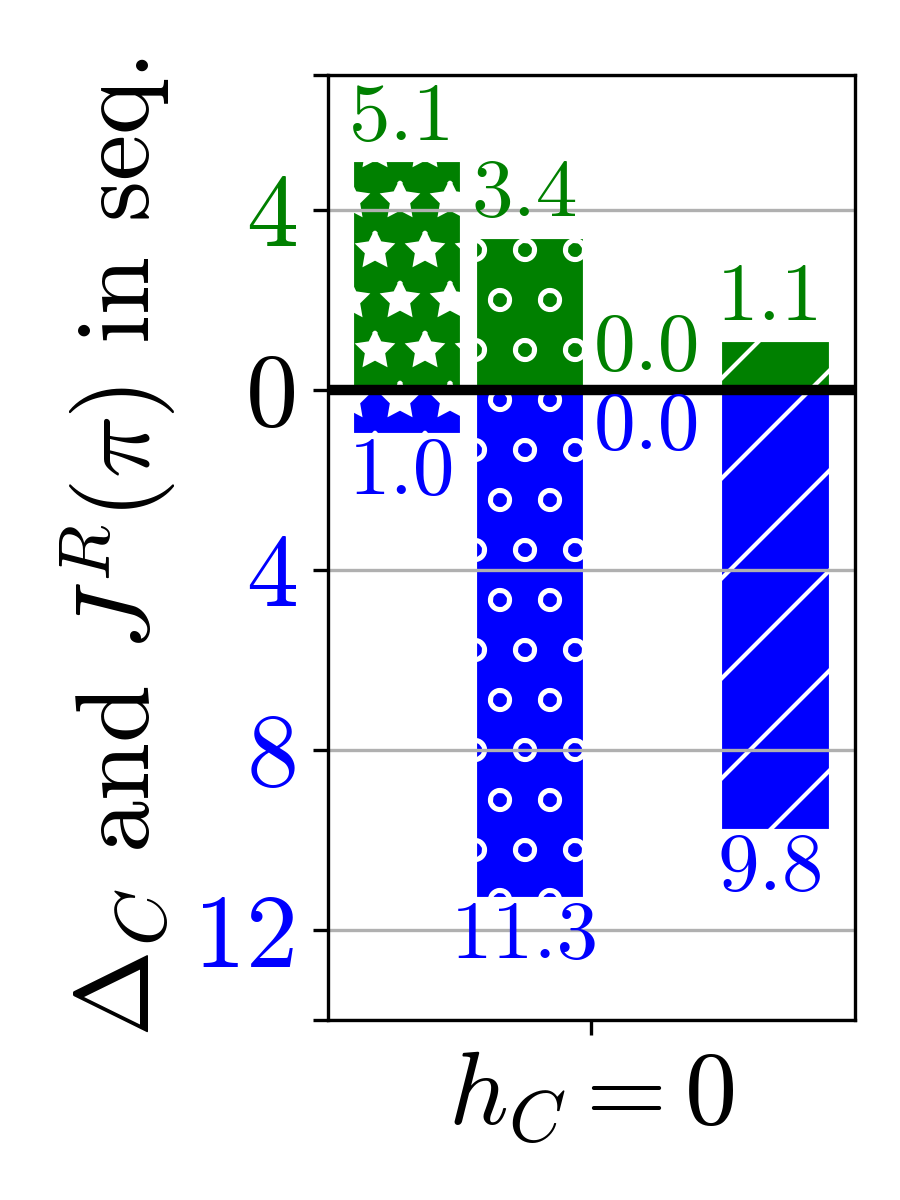}}
%}

\adjustbox{lap={\width}{-1.5mm}}{\raisebox{-0.0mm}{
\includegraphics[width=1.0\linewidth]{figure_result/legend.png}}}

%\vspace{-4mm}

%\subfloat[Ablation studies with $x$ in $\mathcal{D}_\mathrm{eval}$]{
%(b)\raisebox{-2.0mm}{
%\includegraphics[width=0.43\linewidth]{figure_result/ablation.png}}
%}
%\subfloat[Results for the constraint interpreter]{
\iffalse
(c)
\scalebox{0.8}{\raisebox{+13.0mm}{
\setlength\tabcolsep{1.0pt} % default value: 6pt
\begin{tabular}{l*{7}{c}r}
\toprule
 &\multicolumn{2}{c}{Budgetary}
 &\multicolumn{2}{c}{Relational}  
 &\multicolumn{2}{c}{Sequential} \\  \cmidrule(lr){2-3}\cmidrule(lr){4-5}\cmidrule(lr){6-7}
 & Ours  & Rule   
 & Ours  & Rule   
 & Ours  & Rule    \\ \hline
 $M_C$ ACC     & \textbf{96.06}\% & 95.86\% & \textbf{81.46} & 65.49\% & \textbf{91.84}\% &91.36\% \\
$M_C$ AUC       & 98.17\% & - &  90.71 & - & 96.99\% & -\\\hline
$h_C$ MSE       & \textbf{0.21}  & 0.74 & \textbf{0.04} & 0.52 & \textbf{0.0002}   & 0.22  \\
\bottomrule
\end{tabular}}}
%}

\raisebox{+0.0mm}{
\hspace*{-6.5cm}
\includegraphics[width=0.55\linewidth]{figure_result/legend_ablation.png}}
\fi
%\vspace{-2mm}
%
% \vspace{-0.1in}
\caption{
Results in \datasetname-grid for the setting of evaluation with the same reward function as seen in training.
%\textbf{(b)}
%
%Ablations showing the effect of each component in \modelname.
%
%\textbf{(c)} The cell-level accuracy (ACC) and area-under-curve (AUC) of predicting $M_C$, and mean-square-error (MSE) of predicting $h_C$ over three types of constraints on $\mathcal{D}_\mathrm{eval}.$
%
%
\modelname\ achieves higher reward and lower constraint violations over the baselines.
}
\label{table:results}
\vspace{-4mm}
\end{figure*}

Having demonstrated the overall effectiveness of \modelname, our remaining experiments analyze \textbf{(1)} the learned models' performance evaluated on the same reward function as in $\mathcal{D}_\mathrm{train}$, and \textbf{(2)} the importance of each component--$M_B, M_C$ and $h_C$ embedding in \modelname.
For compactness, we restrict our consideration in~\datasetname-grid.

\parab{Evaluation with reward function from $\mathcal{D}_\mathrm{train}$.}
To provide another point of comparison in addition to our main results, we evaluate all models using the same reward function as in $\mathcal{D}_\mathrm{train},$ but with unseen textual constraints from $\mathcal{D}_\mathrm{eval}.$ (
Fig. \ref{table:results})
\begin{wrapfigure}{R}{0.47\textwidth}
\centering
\vspace{-7mm} 
\includegraphics[width=1.0\linewidth]{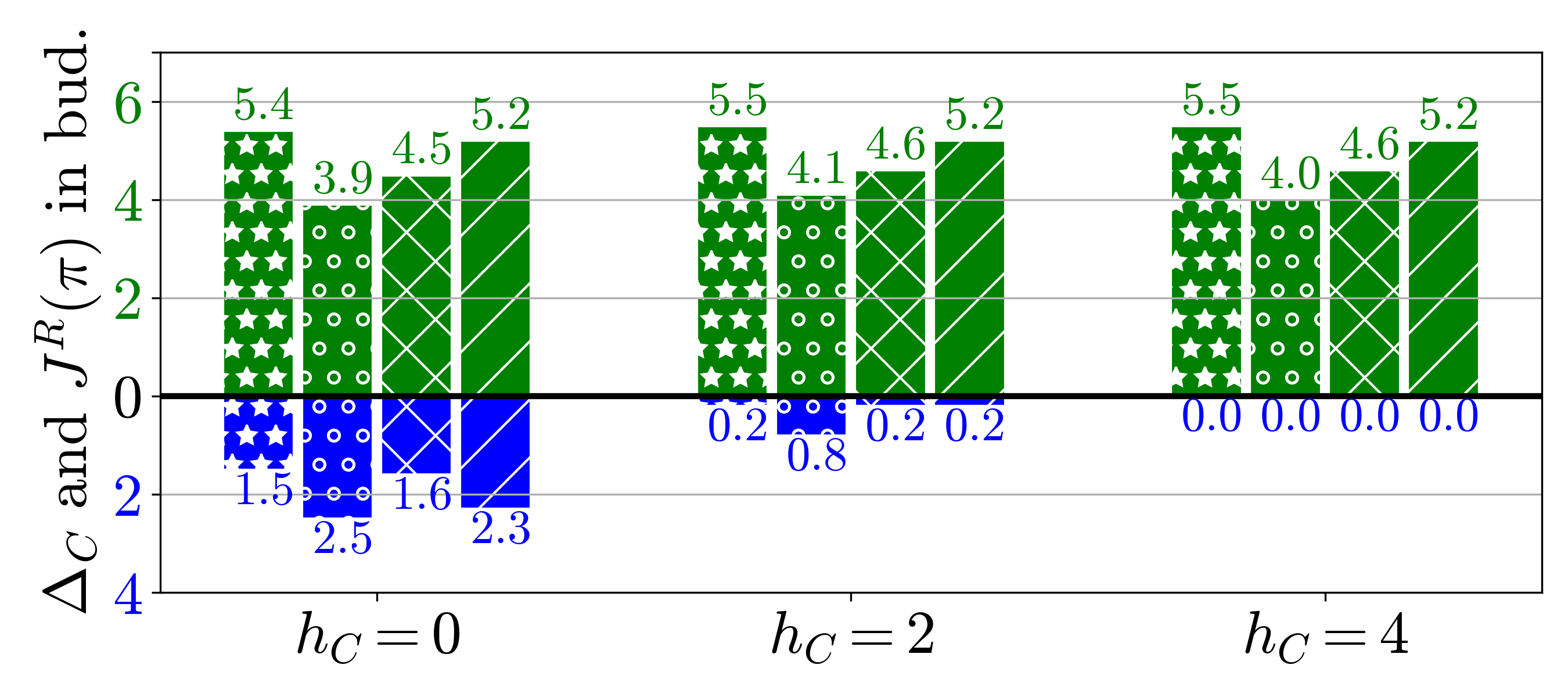}
\raisebox{+0.0mm}{
\hspace*{-6.0mm}
\includegraphics[width=1.05\linewidth]{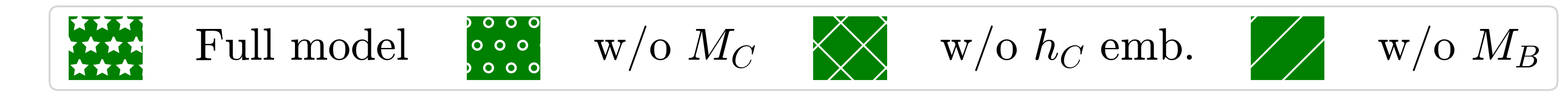}}
% \vspace{-2mm} 
\caption{
Ablations showing the effect of each component in \modelname.
}
\label{fig:abaltion_results}
\vspace{-2mm}
\end{wrapfigure}
%
%Note that all models are trained with language constraints with multiple $h_C$ at the same time and we group the results by $h_C$ for the ease of presentation.
%
We observe \modelname\ achieves the lowest violations across different choices of $h_C$ compared to the baselines.
%For instance, when $h_C=0$ the constraint violation of CF with PCPO is two times larger than that of \modelname.
%
This implies that merely combining the observations and the text is not sufficient to learn an effective representation for parsing the constraints.
%
%This also suggests that our policy network is effective in learning safe policies.
%
%Finally, the PN agent trained with FPO has a low reward, because simply treating the cost penalty as the negative reward hinders the agent's exploration.
%
In addition, \modelname\ achieves the best reward performance under cost satisfaction for the more complex relational and sequential constraints.
For the relational case, 
although the CF agent trained with PCPO satisfies the constraints, it has a relatively low reward.
\parab{Ablation studies.}
We also examine the importance of each part in \modelname{} (Fig.~\ref{fig:abaltion_results}).
To eliminate prediction errors from the constraint interpreter, we use the \textit{true} $M_C$ and $h_C$ here. 
Our full model achieves the best performance in all cases, averaging 5.12\% more reward and 2.22\% fewer constraint violations.
Without $M_C,$ the agent cannot recognize cost entities effectively, which causes the agent to incur 66.67\% higher $\Delta_C$ compared with the full model (which has a $\Delta_C$ close to zero).
This shows that $h_C$ embedding and the $M_B$ mask are useful in enabling constraint satisfaction given textual constraints.

%% file: conclusion.tex
\vspace{-1mm}
\section{Conclusion}
\label{sec:conlusion}
%
% We address the problem of learning safe policies in reifnrocement learning when the constraints are specified by natural language.
% 
% We create a new benchmark task with crowd-sourced text for a variety of constraints.
% %
% We further propose a model and an algorithm to learn constraint-satisfying policies effectively.
% %
% The proposed model achieves superior reward and cost performance compared with other approaches. 
% %
% Future work will consider the following: 
% %
% \textbf{(1)} testing the proposed model given 3D pixel observations,
% %
% \textbf{(2)} learning a good representation that encodes cost information to deal with the combinatorial problem of specifying multiple types of the constraints,
% %
% and \textbf{(3)} considering other types of constraints for safety-critical applications (\eg robotics).

%In this paper, we tackled the problem of safe RL when safety constraints are specified in natural language. 
This work provides a view towards machines that can interoperate with humans. 
As machine agents proliferate into our world, they should understand the safety constraints uttered by the human agents around them. Accordingly, we proposed the problem of safe RL with natural language constraints, created a new benchmark called \datasetname{} to test agents and develop new algorithms for the task, and proposed a new model, \modelname{}, that learns to interpret constraints.

%Our model contains two parts: \textbf{(1)} a constraint interpreter to parse textual constraints into logical form, and \textbf{(2)} a policy network that generates actions based on the constraint interpreter's outputs. 
%We showed how \modelname{} can be trained with existing safe RL algorithms and can learn to safely learn in unseen environments. 
% In experiments, \modelname{} achieves superior reward performance, lower constraint violations, and better generalization compared to several baselines. 
% Relatedly, the goal of \datasetname{} is to create a simulated environment where safe behavior is defined in natural language.

The thesis of POLCO is that modularity enables reuse. By bootstrapping a modular constraint interpreter through exploration, our model scales gracefully to multiple constraints and to shifts in the environment's reward structure, all while exploring new environments safely. We applied POLCO within \datasetname{} to train an agent that navigates safely by obeying natural language constraints. This agent is a step towards creating applications like cleaning robots that can obey free form constraints, such as ``\textit{don't get too close to the TV}''--a relational constraint in our formulation.

% We hope that \datasetname\ and \modelname\ provide a testbed and inspiration to develop new models for safe RL with constraints specified in natural language.
%, with a view towards training any machine task where safety as defined by a human is relevant.

No model is without limitations. The absolute scores on \datasetname{} still leave a lot of room for improvement using better models or training techniques. The current version of \datasetname{} is also not all-encompassing--we envision it as a benchmark that evolves over time, with the addition of new types of constraints and new environments. Future work can investigate training \modelname{} without explicit labels for the constraint interpreter, potentially using techniques like Gumbel softmax \cite{Jang2017CategoricalRW}, or extending \modelname{} to tasks with more realistic visuals.

%% file: appendix.tex
\begin{center}
\Large
\textbf{Supplementary Material}
\end{center}

\parab{Outline.} The supplementary material is outlined as follows.
Section \ref{subsec:appendix_dataset} details the dataset and the procedure of collecting the dataset.
Section \ref{subsec:appendix_parameter} describes the parameters of the constraint interpreter and the policy network, and the PCPO training details.
Section \ref{subsec:appendix_experiment} provides the learning curves of training the policy network.
Section \ref{appendix:robots} details how we apply \modelname\ in the robotics tasks.
Finally, dataset and code to reproduce our experiments are available at~\url{https://sites.google.com/view/polco-hazard-world/}.

%
%Section \ref{subsec:appendix_representation} discusses the possible approaches to learn representations that handles other types of the constraints.
%

\section{Dataset}
\label{subsec:appendix_dataset}

At a high level, \textsc{\datasetname}\ applies the instruction following paradigm to safe reinforcement learning. Concretely, this means that safety constraints in our environment are specified via language. Our dataset is thus comprised of two components: the environment, made up of the objects that the agent interacts with, and the constraint, which imposes a restriction on which environmental states can be visited.

The environment is procedurally generated. For each episode, \textsc{\datasetname}\ places the agent at a randomized start location and fills the environment with objects. \textsc{\datasetname}\ then randomly samples one constraint out of all possible constraints and assigns this constraint to the environment.

\begin{figure*}[t]
\vspace{0mm}
\centering
\subfloat[General prompt for all constraint classes. ]{\includegraphics[width=\linewidth]{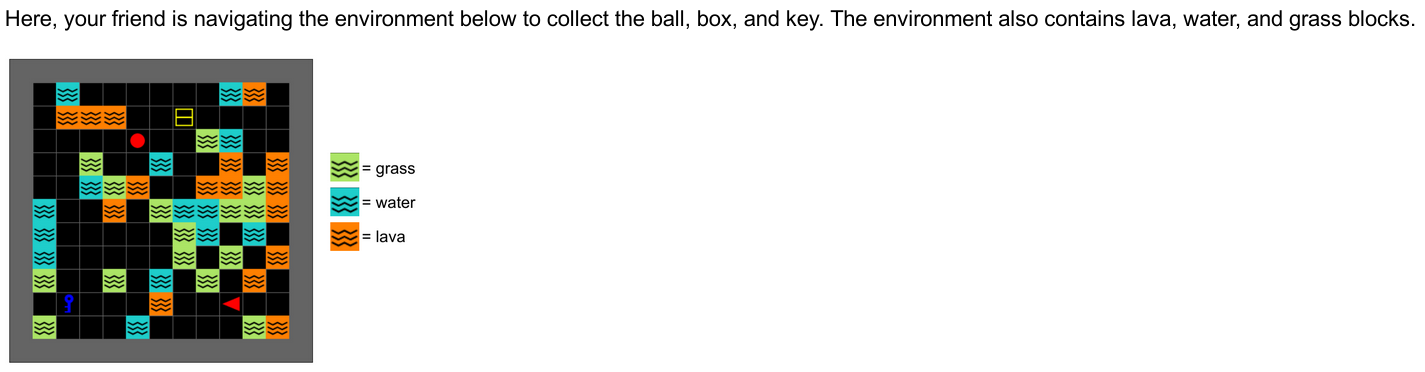}} \\
\subfloat[Budgetary prompt. ]{\includegraphics[width=\linewidth]{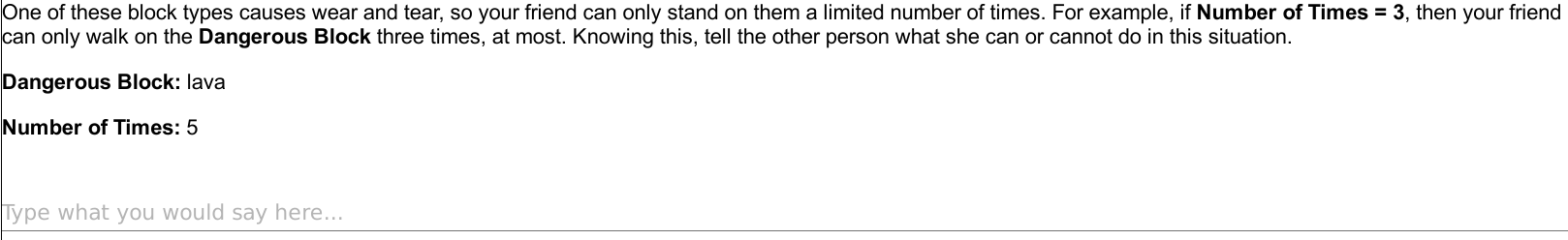}} \\
\subfloat[Relational prompt. ]{\includegraphics[width=\linewidth]{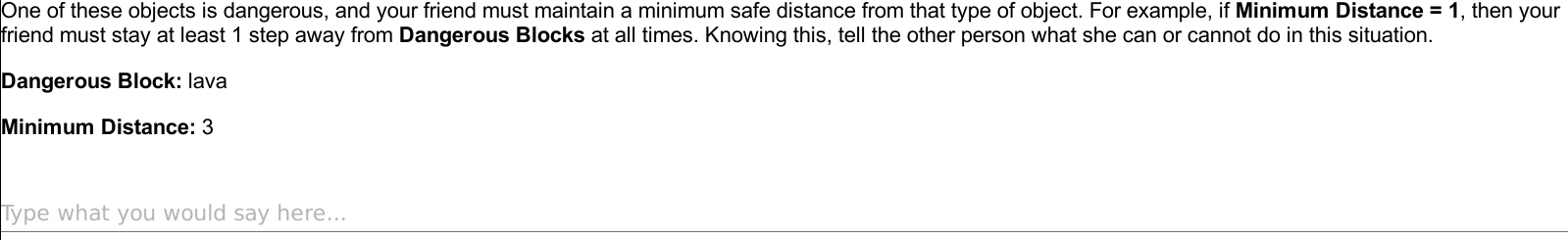}} \\
\subfloat[Sequential prompt. ]{\includegraphics[width=\linewidth]{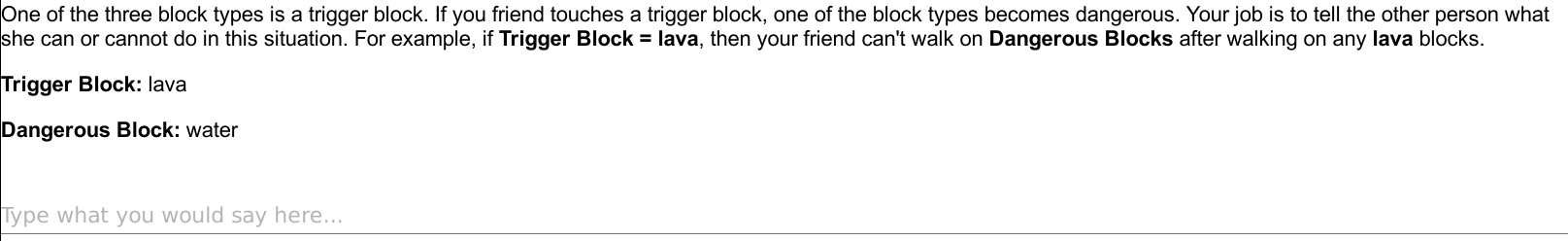}} \\
%\vspace{-2mm}
\caption{
AMT workers receive the general prompt and one of the three specific prompts. They are then asked to instruct another person for the given situation.
This ensures that the texts we collected are free-form. 
}
\label{fig:turk}
\vspace{0mm}
%\end{mdframed}
\end{figure*}

We collected natural language constraints in a two-step process. In the first step, or the data generation step, we prompted workers on Amazon Mechanical Turk with scenarios shown in Fig. \ref{fig:turk}. Workers are provided the minimum necessary information to define the constraint and asked to describe the situation to another person. For example, to generate a so-called budgetary constraint, workers are given the cost entity to avoid (\textit{`lava', `grass'} or \textit{`water'}) and the budget (\ie $h_C,$ a number 0 through 5). The workers use this information to write an instruction for another person.
This allows us to ensure that the texts we collected are free-form.
These generations form our language constraints. 

In the second step, or the data validation step, we employed an undergraduate student to remove invalid constraints. We define a constraint as invalid if (a) the constraint is off-topic or (b) the constraint does not clearly describe states that should be avoided. Examples of valid and invalid constraints are included in Table \ref{tab:valid}. Finally, we randomly split the dataset into 80\% training and 20\% test sets. 
In total, we spent about \$ 1500 for constructing \textsc{\datasetname}.

In \textsc{\datasetname}\ and Lawawall, the agent has 4 actions in total: $a\in\mathcal{A}=\{\mathrm{right},\mathrm{left},\mathrm{up},\mathrm{down}\}.$ 
The transition dynamics $T$ is deterministic.

\begin{table*}[t]
\centering
\vspace{0.0in}
\scalebox{1.0}{
\begin{tabular}{c|p{10cm}}
\toprule
\multicolumn{1}{c}{\textbf{Constraint Type}}
& \multicolumn{1}{c}{\textbf{Examples}} \\ \hline
\multirow{5}{*}{Budgetary} 
& The water should only be stepped on a max of 5 times. \\ \cline{2-2}
& Lava hurts a lot, but you have special shoes that you can use to walk on it, but only up to 5 times, remember! \\ \cline{2-2}
& You can get in lava, but only once. \\ \cline{2-2}
& Four is the most number of times you can touch water \\\cline{2-2}
& You cannot step on the lava block at all. You will die otherwise. \\ \hline
\multirow{5}{*}{Relational} 
& Water will hurt you if you are two steps or less from them. \\ \cline{2-2}
& Always stay 1 step away from lava \\\cline{2-2}
& Any block within one unit of a grass cannot be touched. \\\cline{2-2}
& The explosion radius of grass is three, so stay at least that distance away from grass. \\\cline{2-2}
& Waters are dangerous, so do not cross them. \\
\hline
\multirow{5}{*}{Sequential} 
& Make sure you don't walk on water after walking on grass.  \\ \cline{2-2}
& Do not touch the water or water will become risky. \\ \cline{2-2}
& You may touch the water first, but the lava is dangerous so do not touch it after. \\ \cline{2-2}
& Avoid lava since you can only walk on it once. After that the lava will hurt you. \\ \cline{2-2}
& Water will trigger grass to become dangerous. \\ \hline
\multirow{5}{*}{Invalid}
& good  \\ \cline{2-2}
& move foreward \\ \cline{2-2}
& Just avoid the perimeter when collecting the objects, and you'll be safe. \\ \cline{2-2}
& Your directions are as follows: if you're facing a block with a water block in front of it, walk five blocks ahead \ldots (81 more words)   \\\cline{2-2}

& asdf \\
\bottomrule
\end{tabular}}
\caption{Examples from the various constraint classes. When a constraint does not fully describe all forbidden states in the environment, we classify it as invalid.}
\label{tab:valid}
\end{table*}

\begin{figure*}[t]
\vspace{0mm}
\centering
\includegraphics[width=0.4\linewidth]{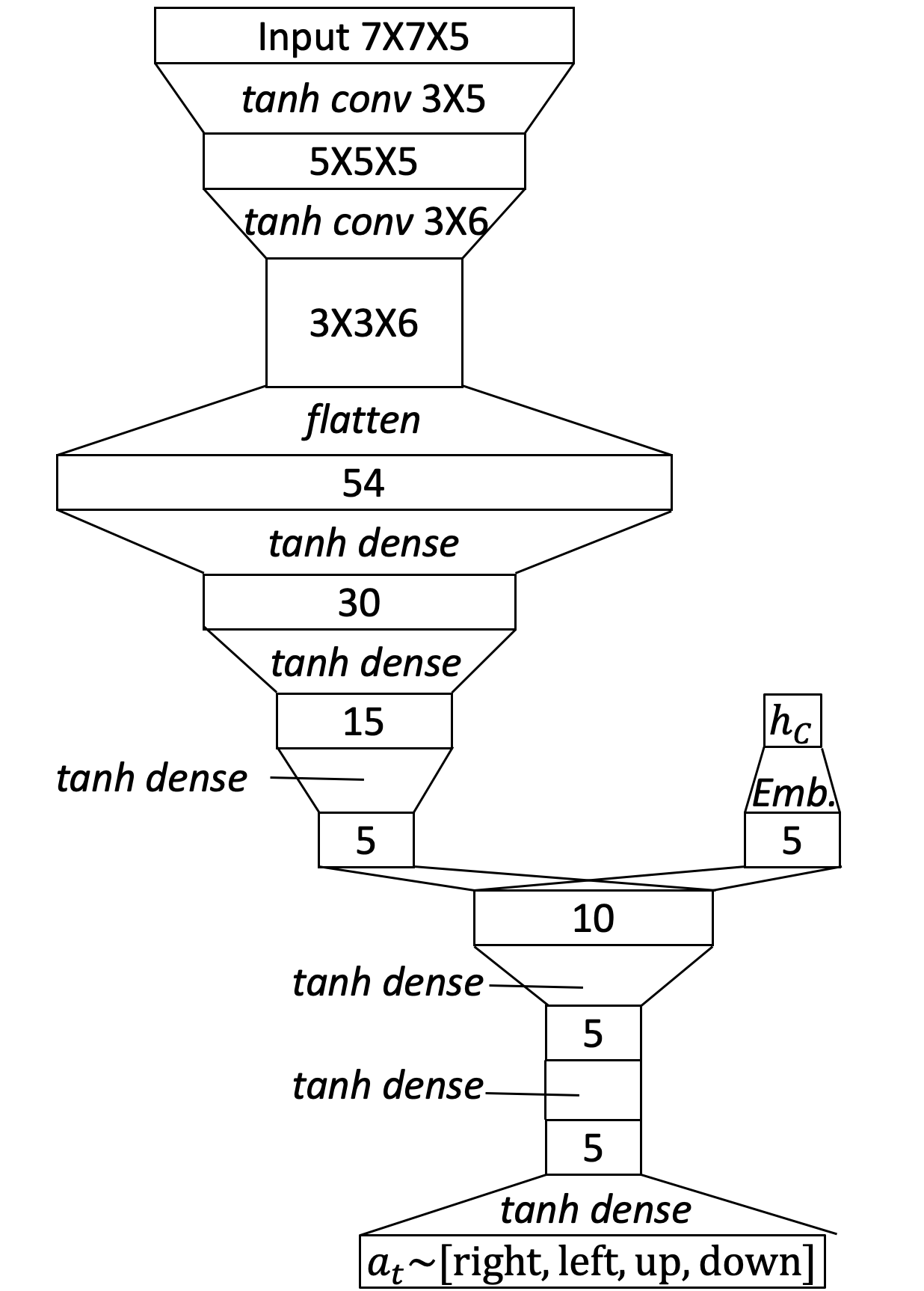}
%\vspace{-2mm}
\caption{
Description of the policy network in \modelname.
}
\label{fig:appendix_model}
\vspace{0mm}
%\end{mdframed}
\end{figure*}

\begin{figure*}[t]
\vspace{0mm}
\centering
\subfloat[Constraint mask module ]{\includegraphics[width=0.45\linewidth]{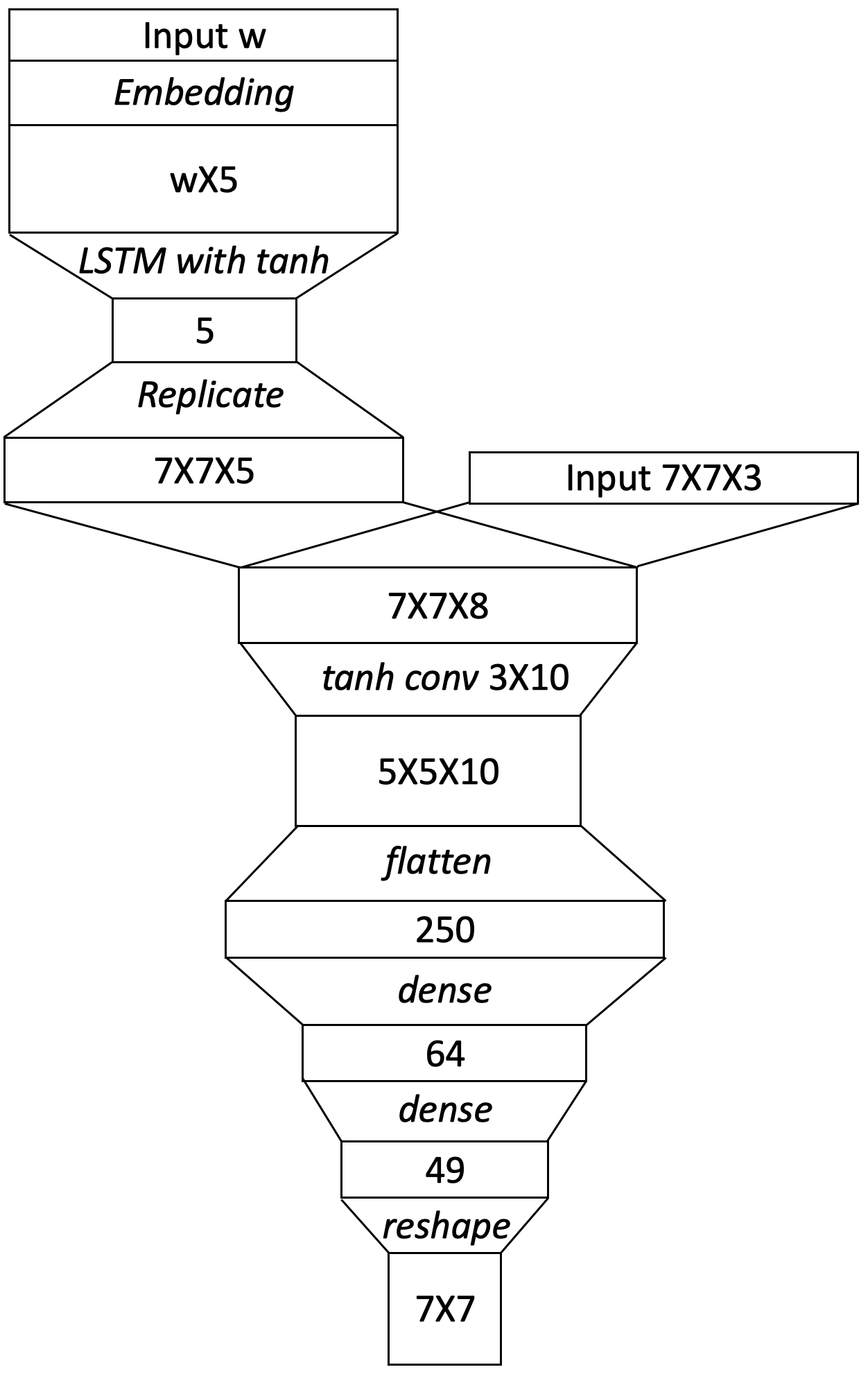}}
\subfloat[Constraint threshold module ]{\includegraphics[width=0.2\linewidth]{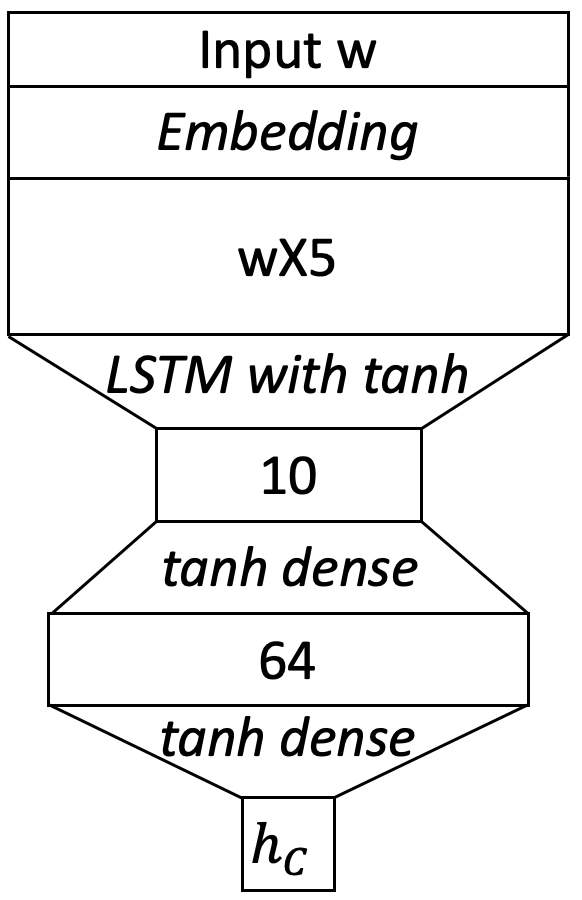}}
%\vspace{-2mm}
\caption{
Description of the constraint interpreter.
}
\label{fig:appendix_model_inter}
\vspace{0mm}
%\end{mdframed}
\end{figure*}

\section{Architectures, Parameters, and Training Details}
\label{subsec:appendix_parameter}
\parab{Policy network in \modelname.} 
The architecture of the policy network is shown in Fig.~\ref{fig:appendix_model}.
The environment embedding for the observation $o_t$ is of the size 7$\times$7$\times$3.
This embedding is further concatenated with the cost constraint mask $M_C$ and the cost budget mask $M_B$.
This forms the input with the size 7$\times$7$\times$5.
We then use convolutions, followed by dense layers to get a vector with the size 5.
This vector is further concatenated with the $h_C$ embedding.
Finally, we use dense layers to the categorical distribution with four classes (\ie turn right, left, up or down in \textsc{\datasetname}).
We then sample an action from this distribution.

\parab{Constraint interpreter in \modelname.} The architecture of the constraint interpreter is shown in Fig.~\ref{fig:appendix_model_inter}.
For the constraint mask module, 
the input is the text with $w$ words.
We then use an embedding network, followed by an LSTM to obtain the text embedding with the size 5.
The text embedding is duplicated to get a tensor with the size 7$\times$7$\times$5.
This tensor is concatenated with the observation of size $7\times7\times3$, creating a tensor with the size 7$\times$7$\times$8.
In addition, we use a convolution, followed by dense layers and a reshaping to get the cost constraint mask $M_C.$ 
Next, we use a heuristic to compute $\hat{C}_{tot} := \sum_{t=0}^{t'}C(s_t,a_t;x)$ from $M_C$.
At execution time, we give our constraint interpreter access to the agent's actions. We initialize $\hat{C}_{tot} =0.$
Per timestep, our agent either turns or moves forward. If the agent moves forward and the square in front of the agent contains a cost entity according to $M_C$, we increment $\hat{C}_{tot}$.
For the constraint threshold module, we use the same architecture to get the text embedding.
We then use dense layers to predict the value of $h_C.$

\begin{table*}[t]
\centering
\vspace{0.0in}
\scalebox{1.0}{
\begin{tabular}{cc}
\toprule
Parameter                                      &  \\  \hline
\multirow{1}{*}{Reward dis. factor~$\gamma$}      & 0.99  \\
\multirow{1}{*}{Constraint cost dis. factor~$\gamma_{C}$}      & 1.0   \\
\multirow{1}{*}{step size~$\delta$}             & $10^{-3}$   \\
\multirow{1}{*}{$\lambda^\mathrm{GAE}_{R}$}    & 0.95  \\
\multirow{1}{*}{$\lambda^\mathrm{GAE}_{C}$}    & 0.9  \\
\multirow{1}{*}{Batch size}                    & 10,000  \\
\multirow{1}{*}{Rollout length}                & 200  \\
\multirow{1}{*}{Number of policy updates} & 2,500 \\
\bottomrule
\end{tabular}}
\caption{Parameters used in \modelname.}
\label{tab:parab}
\end{table*}

\iffalse
\parab{Rule-based baseline.}
%
The rule-based baseline is similar to a $n$-gram model, in which we compute the likelihood of the cost entity given the text.
%
Specifically, in the training set we count the number of \textit{word-cost entity} pairs. 
%
For example, for each map, given the language constraints ``\textit{The water should only be stepped on a max of 5 times}'' and the ground-truth cost entity (\eg `\textit{water}'), the pair `\textit{the}'-`\textit{water}' is added one and so forth.
%
During testing, for a given text and map, we compute the following probability for each possible cost entity:
\begin{align}
    P(`\textit{water}'|\text{textual constraints})\propto &P(\text{textual constraints})P(`\textit{water}') \nonumber\\
    = & \Pi_{\text{word}} P(\text{word}|`\textit{water}')P(`\textit{water}')\nonumber,
\end{align}

\begin{align}
    P(`\textit{lava}'|\text{textual constraints})\propto &P(\text{textual constraints})P(`\textit{lava}') \nonumber\\
    = & \Pi_{\text{word}} P(\text{word}|`\textit{lava}')P(`\textit{lava}')\nonumber,
\end{align}

\begin{align}
    P(`\textit{grass}'|\text{textual constraints})\propto &P(\text{textual constraints})P(`\textit{grass}') \nonumber\\
    = & \Pi_{\text{word}} P(\text{word}|`\textit{grass}')P(`\textit{grass}')\nonumber.
\end{align}
Finally, we select the one with the maximum value to predict the cost entity.
\fi
%
\parab{Details of the algorithm--PCPO.}
We use a KL divergence projection in PCPO to project the policy onto the cost constraint set since it has a better performance than $L_2$ norm projection.
We use GAE-$\lambda$ approach \cite{schulman2015high} to estimate $A^\pi_{R}(s,a)$ and $A^\pi_{C}(s,a).$
We use neural network baselines with the same architecture and activation functions as the policy networks.
The hyperparameters of training \modelname\ are in Table \ref{tab:parab}. 
We conduct the experiments on the machine with Intel Core i7-4770HQ CPU.
The experiments are implemented in rllab~\cite{duan2016benchmarking}, 
a tool for developing RL algorithms.

\begin{figure*}[t]
\vspace{0mm}
\centering
\includegraphics[width=0.4\linewidth]{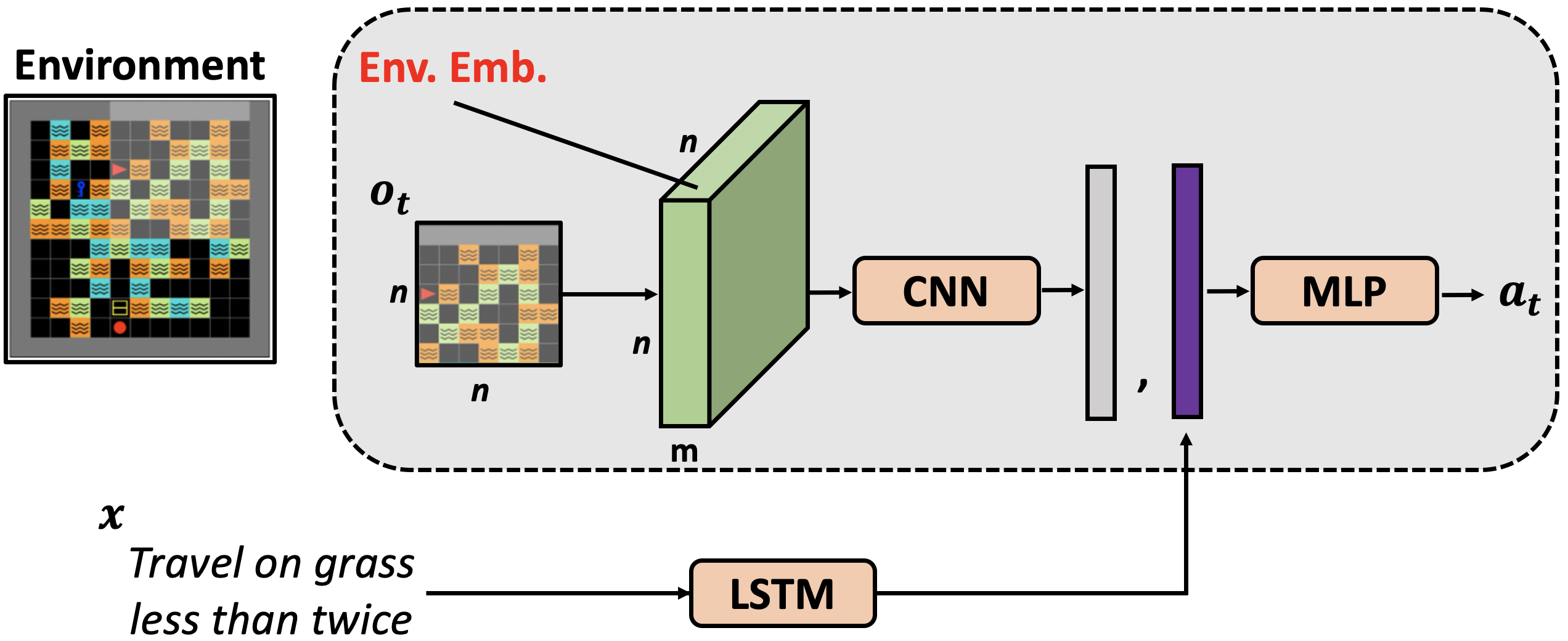}
%\vspace{-2mm}
\caption{
Baseline model--Constraint Fusion (CF).
It is composed of two parts -- \textbf{(1)} a CNN takes $o_t$ as an input and produce a vector representation,
\textbf{(2)} an LSTM takes $x$ as an input and produce a vector representation.
We then concatenate these two vectors, followed by a MLP to produce an action $a_t.$
}
\label{fig:appendix_baseline}
\vspace{0mm}
%\end{mdframed}
\end{figure*}

\begin{figure*}[t]
\vspace{0mm}
\centering
\includegraphics[width=0.5\linewidth]{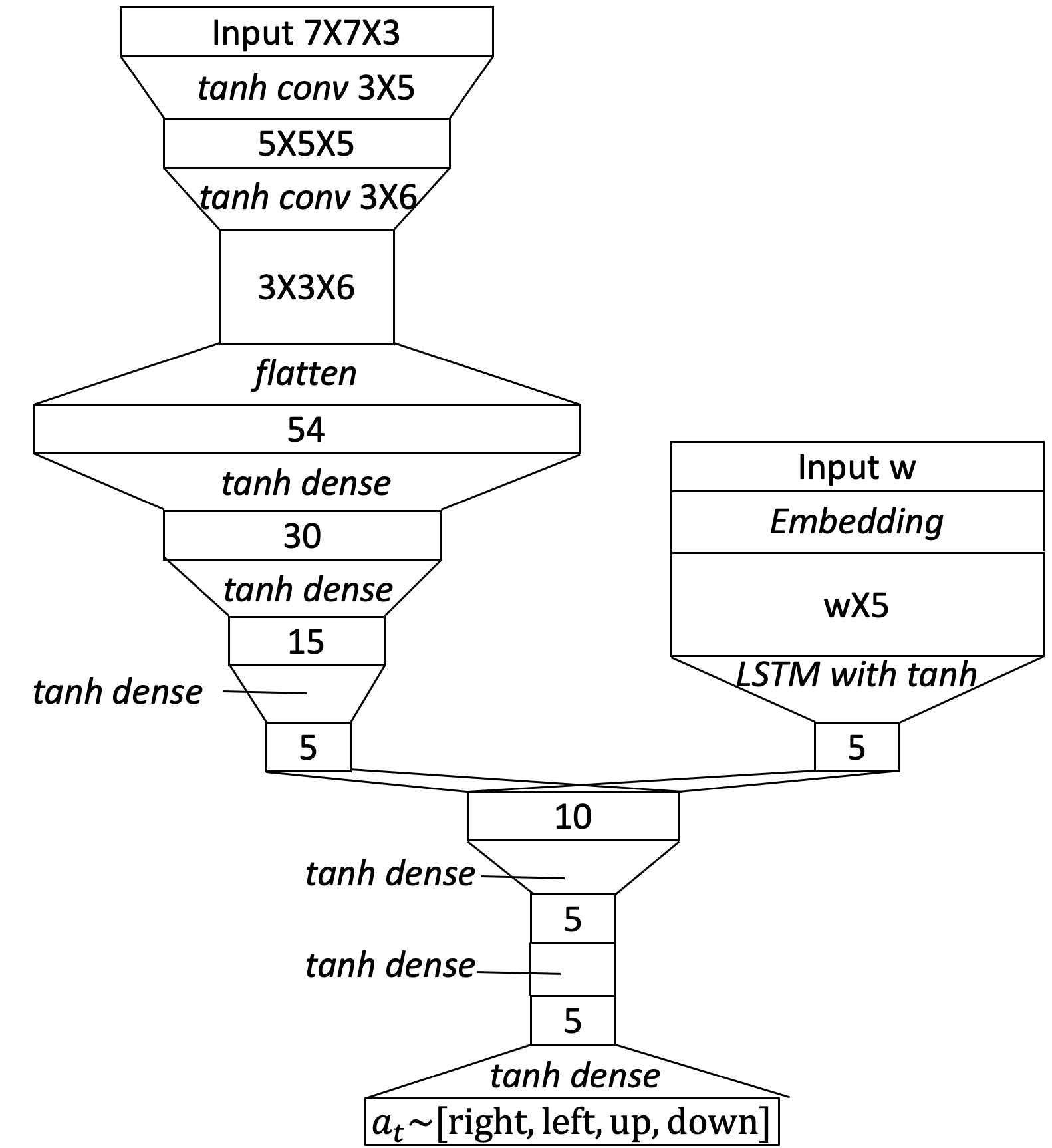}
%\vspace{-2mm}
\caption{
Description of our baseline model-Constraint Fusion (CF).
}
\label{fig:appendix_model_baseline_parameter}
\vspace{0mm}
%\end{mdframed}
\end{figure*}

\parab{Baseline model--Constraint Fusion (CF).}
The model is illustrated in Fig. \ref{fig:appendix_baseline}.
An LSTM takes the text $x$ as an input and produces a vector representation.
The CNN takes the environment embedding of $o_t$ as an input and produces a vector representation.
These two vector representations are concatenated, followed by a MLP to produce an action $a_t.$
We do not consider other baselines in \cite{janner2017representation} and \cite{misra2018mapping}.
This is because that their models are designed to learn a multi-modal representation (\eg processing a 3D vision) and follow goal instructions.
In contrast, our work focuses on learning a constraint-satisfying policy. 
The parameters of the baseline is shown in Fig. \ref{fig:appendix_model_baseline_parameter}.
We use the same CNN parameters as in our policy network to process $o_t.$
Then, we use the same LSTM parameters as in our constraint mask module to get a vector representation with size 5.
Note that we use almost the same number of the parameters to ensure that \modelname\ does not have an advantage over CF.
Finally, we use dense layers to the categorical distribution with four classes. 
We then sample an action from this distribution.

\section{Additional Experiments}
\label{subsec:appendix_experiment}

\begin{figure*}[t]
%\vspace{-1mm}
\centering
\subfloat[Budgetary constraints]{
\includegraphics[width=0.33\linewidth]{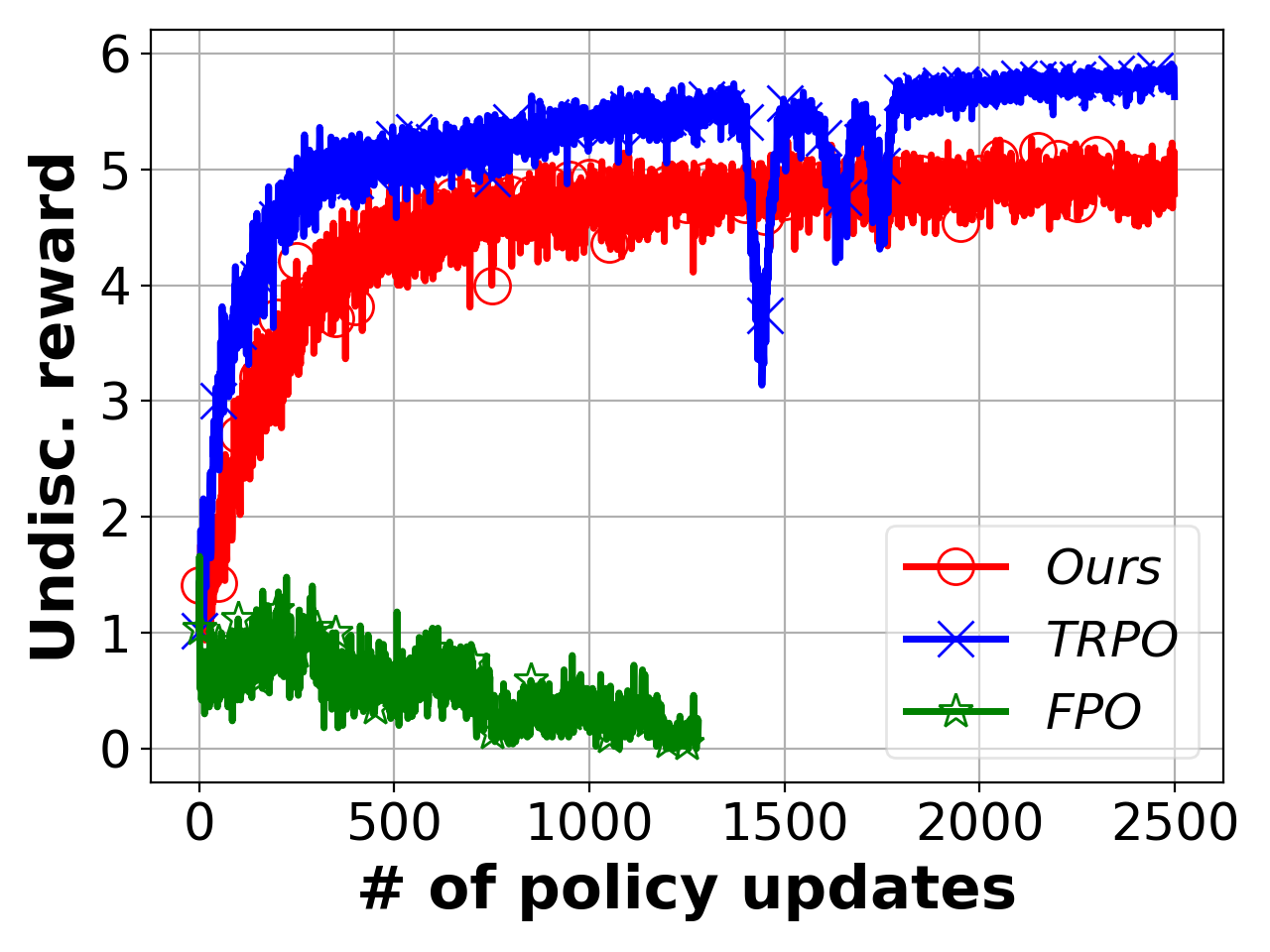}
\includegraphics[width=0.33\linewidth]{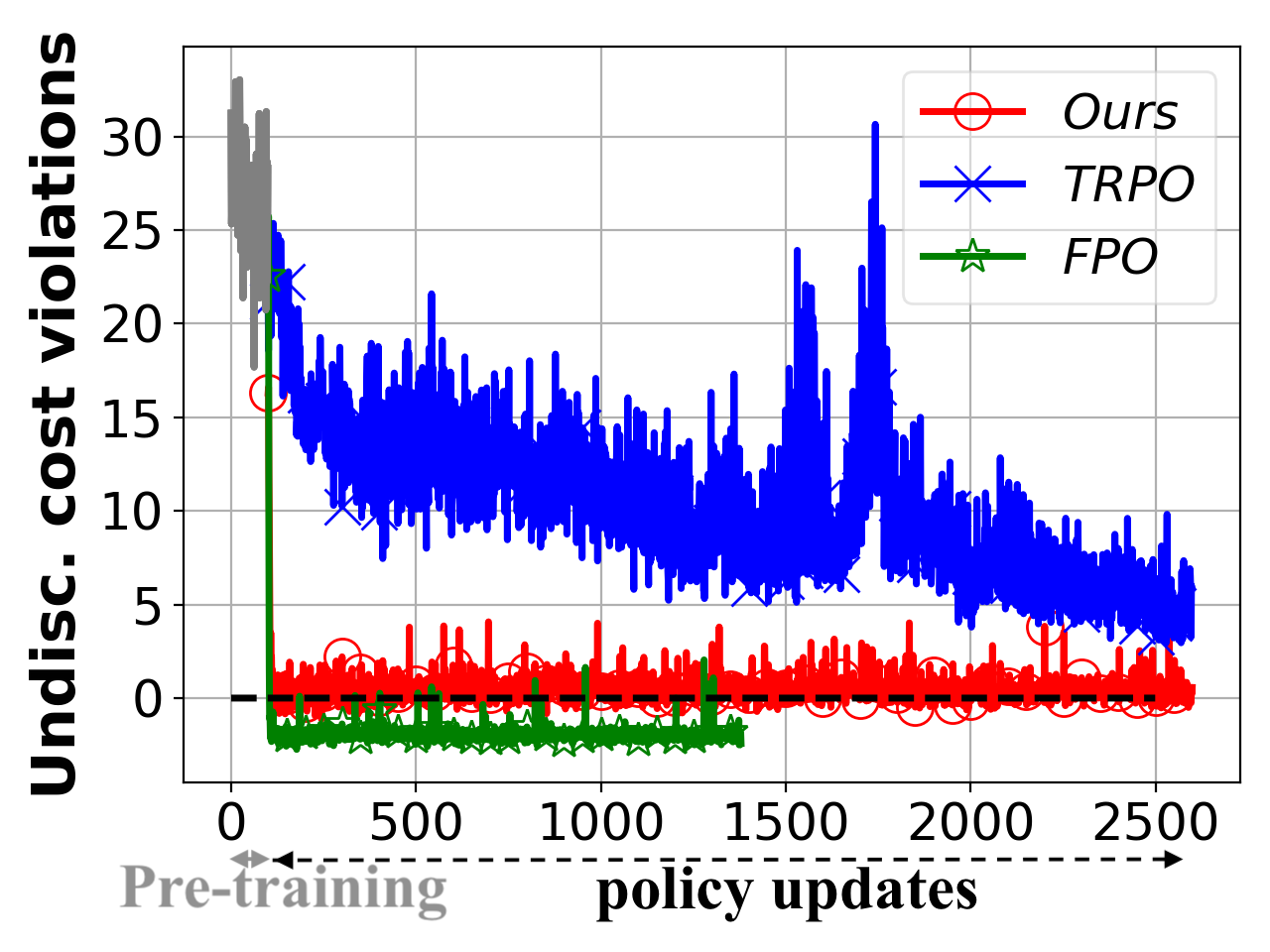}
\includegraphics[width=0.33\linewidth]{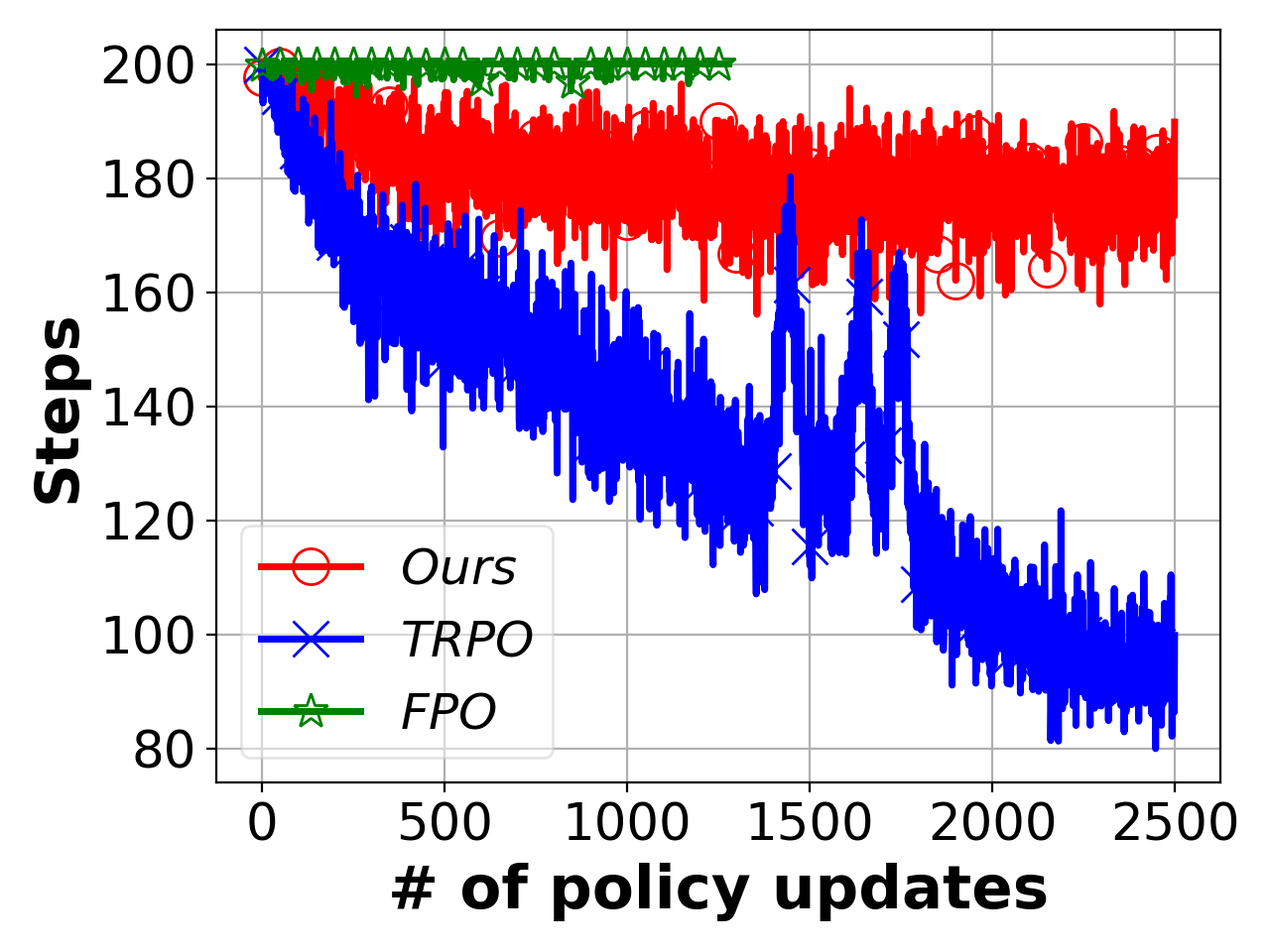}}

\subfloat[Relational constraints]{
\includegraphics[width=0.33\linewidth]{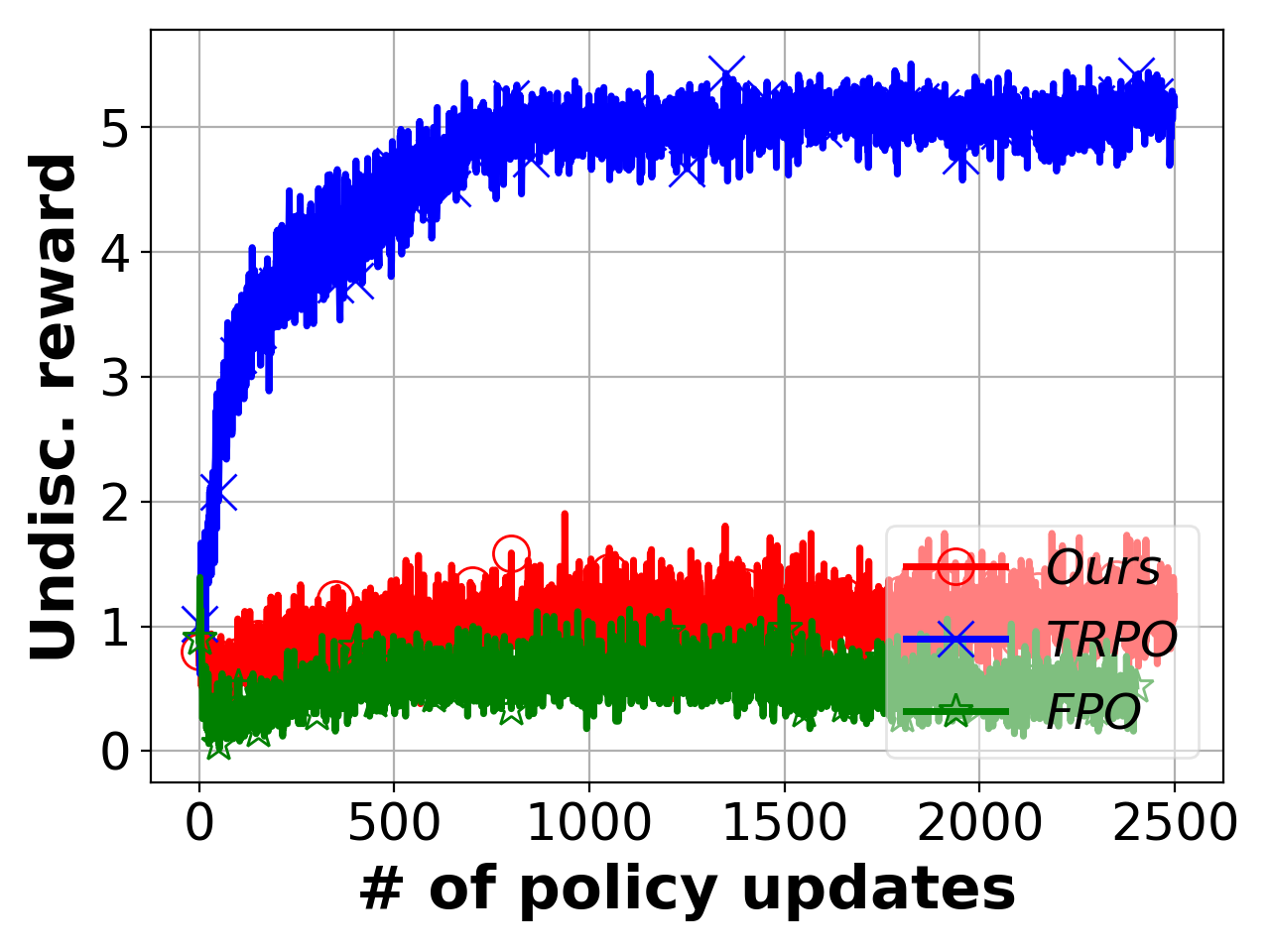}
\includegraphics[width=0.33\linewidth]{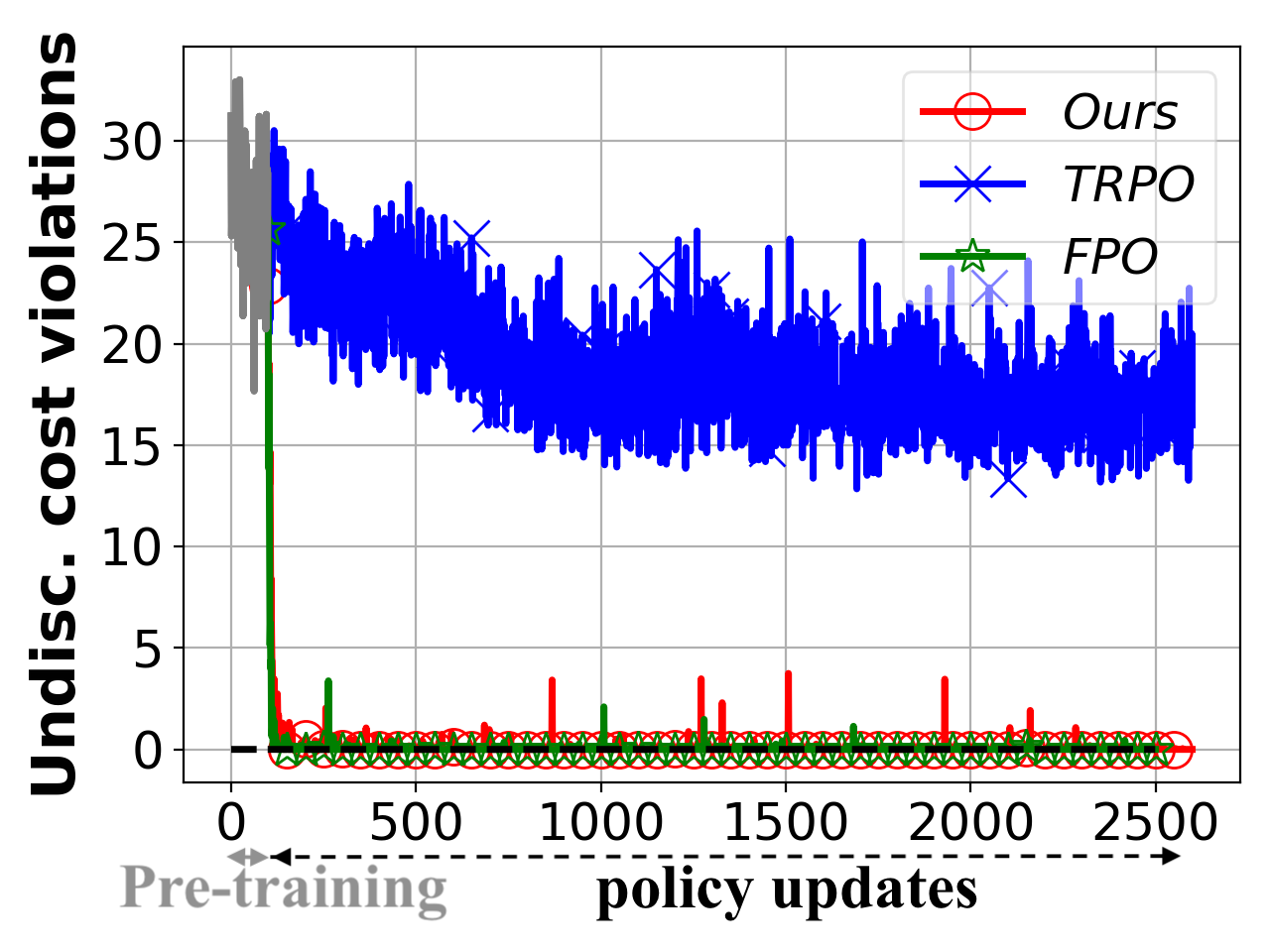}
\includegraphics[width=0.33\linewidth]{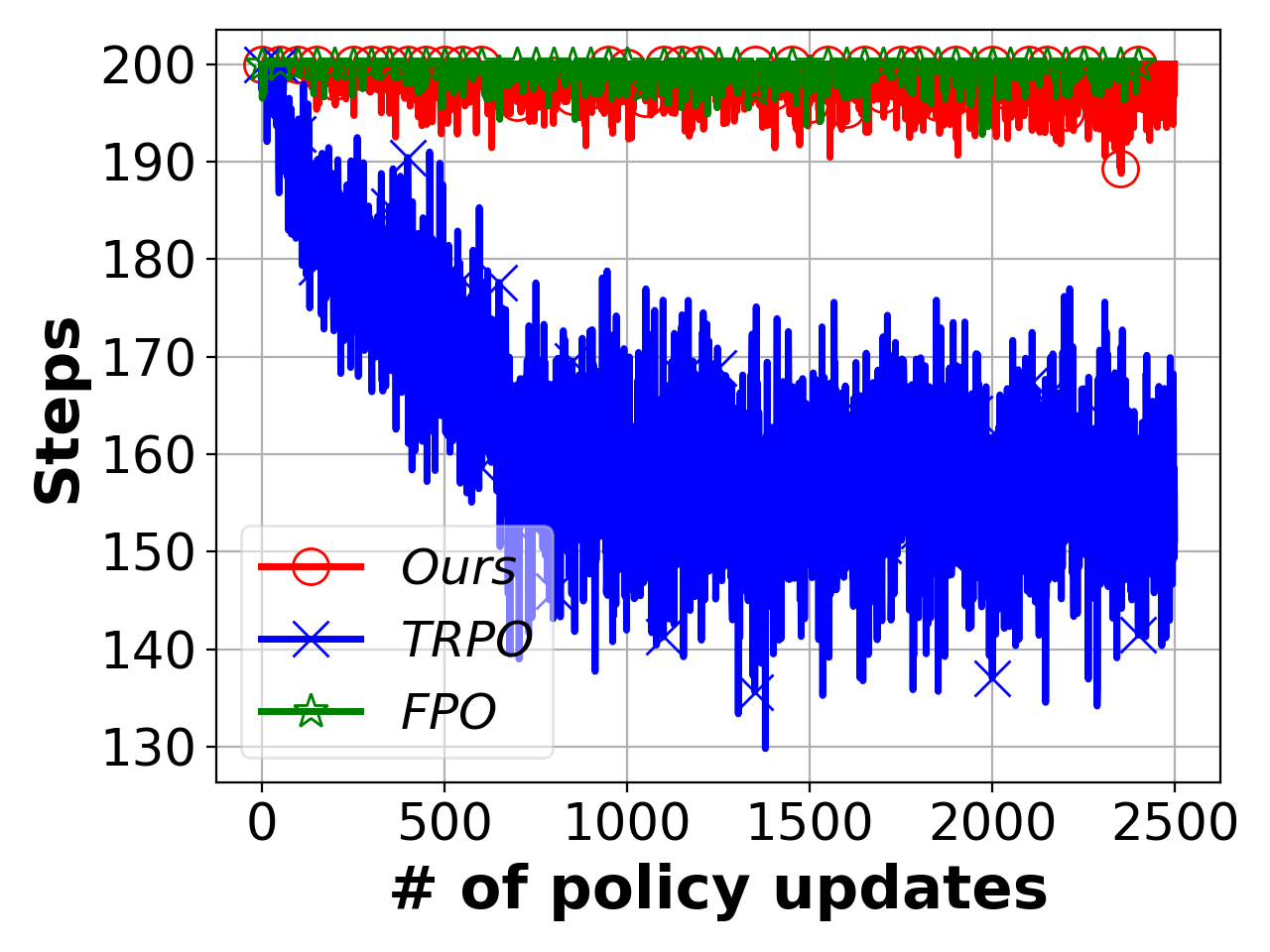}}

\subfloat[Sequential constraints]{
\includegraphics[width=0.33\linewidth]{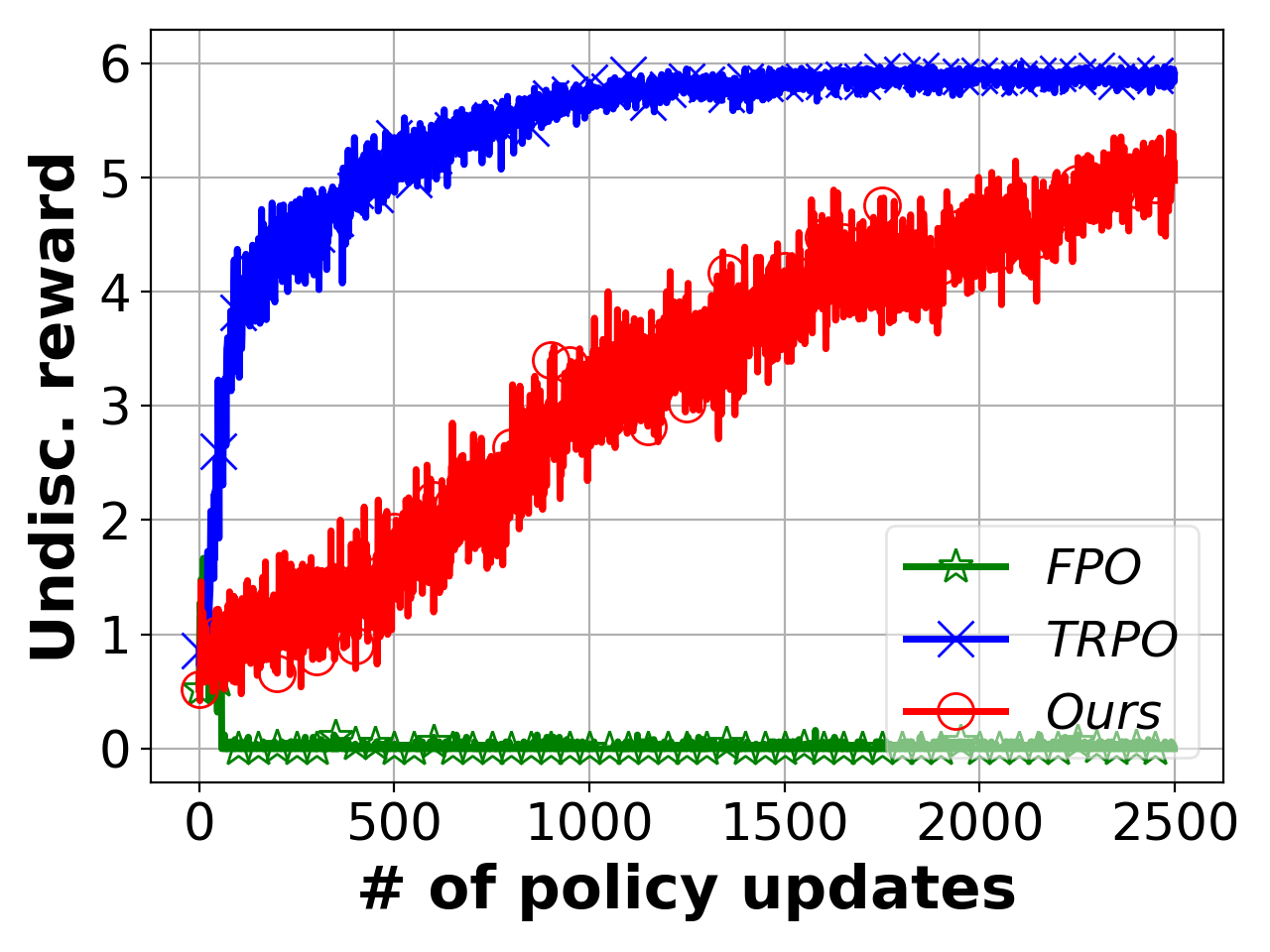}
\includegraphics[width=0.33\linewidth]{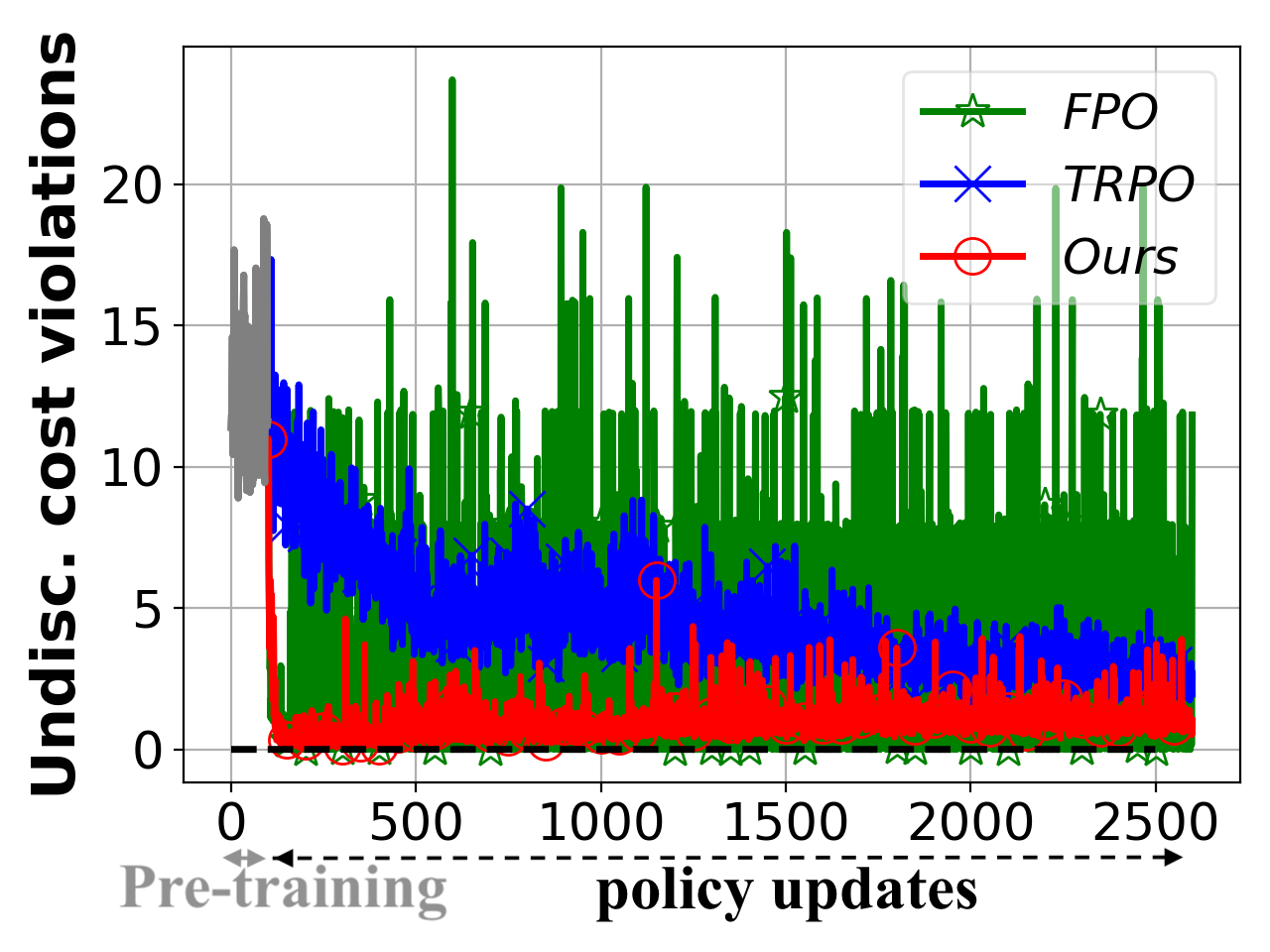}
\includegraphics[width=0.33\linewidth]{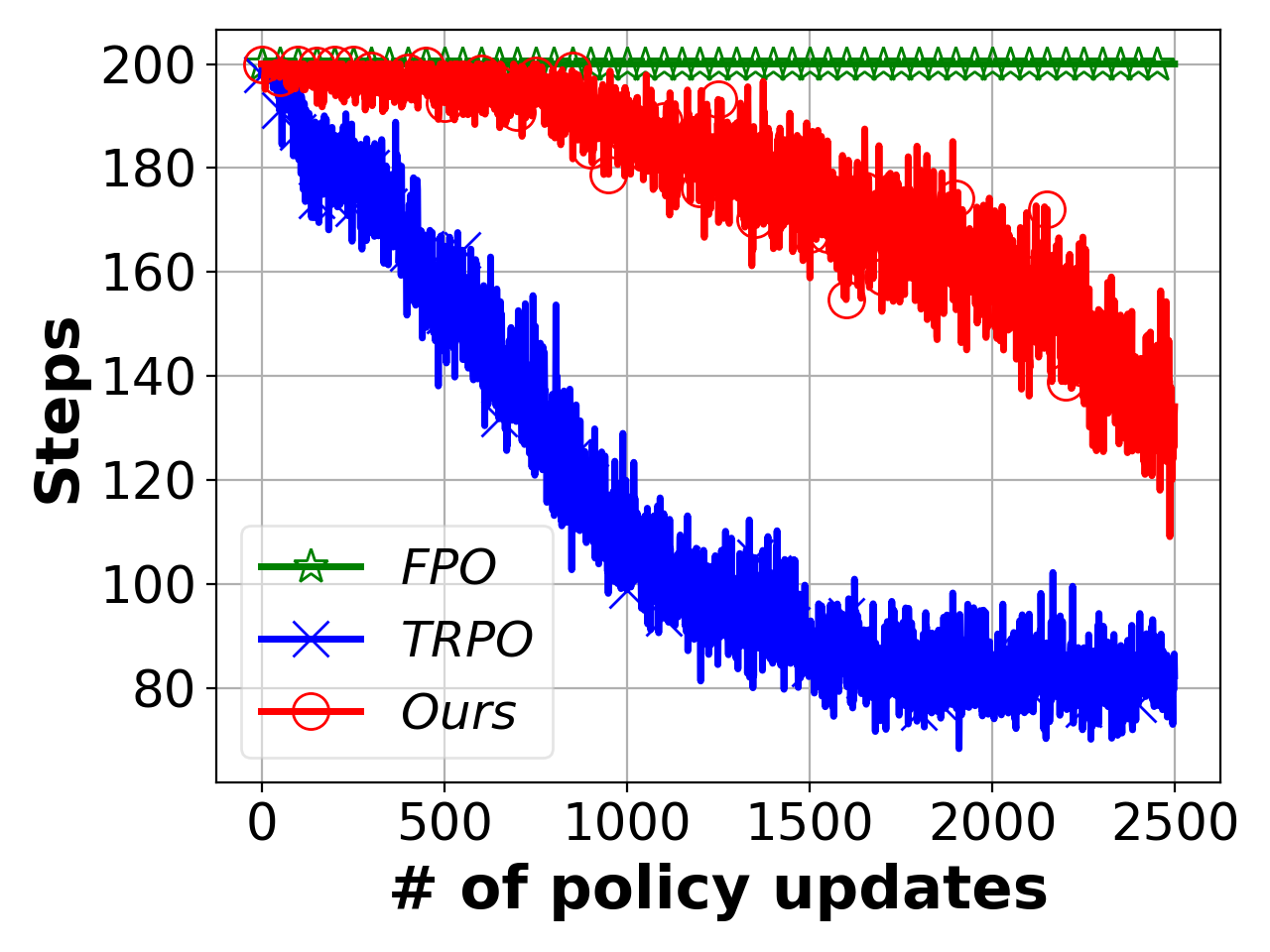}}
%\vspace{-2mm}
\caption{
\textbf{Learning curves of training the policy network.}
The undiscounted reward, the undiscounted cost violations (\ie $\Delta_C = J_C(\pi)-h_C$), and the number of steps over policy updates for the tested algorithms and the constrains.
In the undiscounted cost violations plots, we further include the numbers for the interpreter pre-training stage in the first 100 points. This is equal to 5000 trajectories.
The maximum allowable step for each trajectory is 200.
We observe that \modelname\ satisfies the cost constraints throughout training while improving the reward.
In contrast, the policy network trained with TRPO suffers from violating the constraints and the one trained with FPO cannot effectively improve the reward.
%
%{\color{}In addition, the cost violations at the pre-training stage is quite small overall considering the x-axis runs to 2500 points at the policy update stage in the figure.}
%
(Best viewed in color.)
}
\label{fig:appendix_learningCurve}
%\vspace{-6.1mm}
%\end{mdframed}
\end{figure*}

\parab{Learning curves of training the policy network.}~
The learning curves of the undiscounted constraint cost, the discounted reward, and the number of steps over policy updates are shown for all tested algorithms and the constrains in Fig. \ref{fig:appendix_learningCurve}.
Overall, we observe that \\
\begin{itemize}
    \item[(1)] \modelname\ improves the reward performance while satisfying the cost constraints during training in all cases,
    \item[(2)] the policy network trained with TRPO has substantial cost constraint violations during training,
    \item [(3)] the policy network trained with FPO is overly restricted, hindering the reward improvement.
\end{itemize}

\section{\modelname\ for 3D robotics tasks.}
\label{appendix:robots}

\begin{figure*}[t]
\vspace{0mm}
\centering
\includegraphics[width=0.5\linewidth]{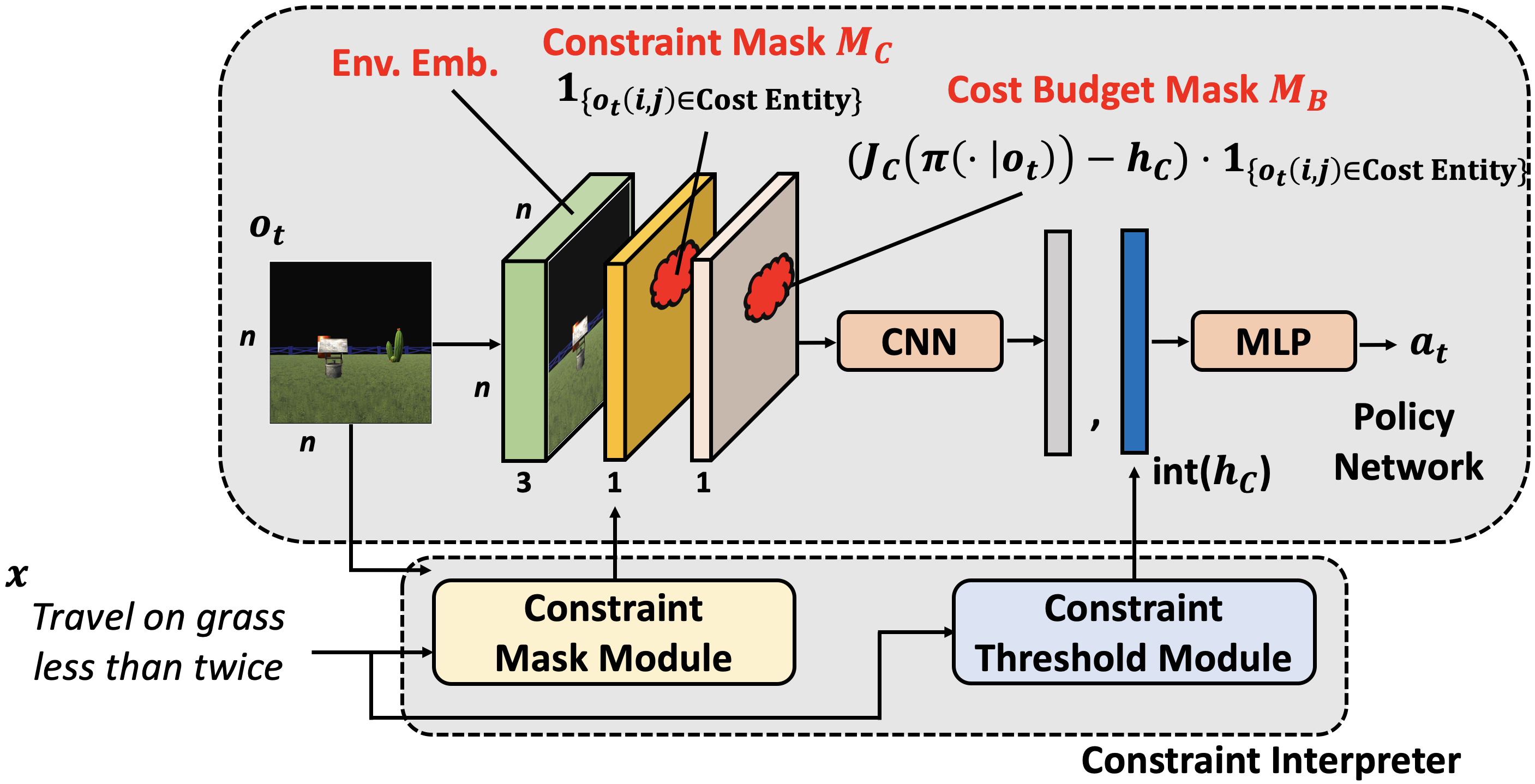}
%\vspace{-2mm}
\caption{
\modelname\ for pixel observations and 3D ego-centric observations. The red cloud area represents the bounding box of each object in $o_t.$ 
}
\label{fig:appendix_model_image}
\vspace{0mm}
%\end{mdframed}
\end{figure*}

To deal with pixel observations $o_t,$ we can still use the proposed architecture to process $o_t$ as shown in Fig. \ref{fig:appendix_model_image}.
To predict the cost constraint mask $\hat{M}_C,$ we use the object segmentation method to get the bounding box of each object in the scene.
As a result, the area of that bounding box will be one if there is a cost entity (\ie the forbidden states mentioned in the text). Otherwise, the bounding box contains a zero.
For $\hat{M}_B,$ we can use a similar approach to compute the cumulative cost violations at each step.
In addition, to deal with navigation environments with 3D ego-centric observations, we propose shifting the $o_t,$ $\hat{M}_C$ and $\hat{M}_B$ matrices to be the first-person view.
The bounding box approach for image case can still be applied here.
We leave this proposal to future work.
